%% file: main.tex
\documentclass{article}

     \usepackage[preprint,nonatbib]{neurips_2020}

\usepackage[utf8]{inputenc} 
\usepackage[T1]{fontenc}    
\usepackage{booktabs}       
\usepackage{amsfonts}       
\usepackage{nicefrac}       
\usepackage{microtype}      

\usepackage{geometry}
\PassOptionsToPackage{hyphens}{url}
\usepackage{hyperref}
\usepackage{graphicx}
\usepackage{subfigure}
\usepackage{enumitem}
\usepackage{url}            
\usepackage{braket} 
\usepackage{amsmath}
\usepackage{color}
\usepackage{mathtools}
\usepackage[toc,page]{appendix}
\usepackage{amsthm}
\usepackage{wrapfig}
\usepackage{float}
\usepackage{tcolorbox}
\usepackage{thmtools}
\usepackage{thm-restate}
\newtheorem{theorem}{Theorem}

\usepackage{algpseudocode}
\usepackage[ruled,vlined,linesnumbered]{algorithm2e}

\usepackage{listings}
\usepackage{xcolor} 

\tcbset{boxsep=0mm,boxrule=0pt,colframe=white,arc=0mm,left=0.5mm,right=0.5mm}

\title{Bloom Origami Assays: Practical Group Testing}

\author{
  Louis Abraham \\
  ETH Zurich\\
  Zurich, Switzerland\\
  \texttt{louis.abraham@yahoo.fr} \\
   \And
   Gary B\'{e}cigneul \\
   ETH Z\"urich\\
   Z\"urich, Switzerland\\
   \texttt{gary.becigneul@inf.ethz.ch} \\
   \And
   Benjamin Coleman \\
   Rice University \\
   Houston, TX \\
   \texttt{ben.coleman@rice.edu} \\
   \AND
   Bernhard Sch\"{o}lkopf \\
   MPI for Intelligent Systems \\
   T\"{u}bingen, Germany \\
   \texttt{bs@tuebingen.mpg.de} \\
   \And
   Anshumali Shrivastava \\
   Rice University \\
   Houston, TX \\
   \texttt{anshumali@rice.edu} \\
   \And
   Alexander Smola \\
   Amazon Web Services\\
   Palo Alto, CA\\
   \texttt{smola@amazon.com} \\
}

  \newcommand{\garyb}[1]{{\color{red}~Gary: #1}}
    \newcommand{\louis}[1]{{\color{purple}~Louis: #1}}

\urldef\myurlevalmode\url{https://louisabraham.github.io/crackovid/crackovid.html?input=MyAzCgowLjk5IDAuOTUKCjAuMSAwLjEgMC4xIAoKZXZhbAoKMDExCjEwMQoxMTAKCjAwMA%3D%3D}

\urldef\myurloptimmode\url{https://louisabraham.github.io/crackovid/crackovid.html?input=MyAzCgowLjk5IDAuOTUKCjAuMSAwLjEgMC4xCgpvcHRpbSBjb25maWRlbmNlCmdhLWx1YnkgMiAxMDAKMTAwMA%3D%3D}

\begin{document}

\maketitle


\begin{abstract}
    We study the problem usually referred to as \textit{group testing} in the context of COVID-19. Given $n$ samples collected from patients, how should we select and test mixtures of samples to maximize information and minimize the number of tests? Group testing is a well-studied problem with several appealing solutions, but recent biological studies impose practical constraints for COVID-19 that are incompatible with traditional methods. Furthermore, existing methods use unnecessarily restrictive solutions, which were devised for settings with more memory and compute constraints than the problem at hand. This results in poor utility. In the new setting, we obtain strong solutions for small values of $n$ using evolutionary strategies. We then develop a new method combining Bloom filters with belief propagation to scale to larger values of $n$ (more than 100) with good empirical results. We also present a more accurate decoding algorithm that is tailored for specific COVID-19 settings. This work demonstrates the practical gap between dedicated algorithms and well-known generic solutions. 
    Our efforts results in a new and practical multiplex method yielding strong empirical performance without mixing more than a chosen number of patients into the same probe. Finally, we briefly discuss adaptive methods, casting them into the framework of adaptive sub-modularity. 
\end{abstract}


\section{Introduction}
\vglue-2mm
Lacking effective treatments or vaccinations, the most effective way to save lives in an ongoing epidemic is to mitigate and control its spread. This can be done by testing and isolating positive cases early enough to prevent subsequent infections. If done regularly and for a sufficiently large fraction of susceptible individuals, mass testing has the potential to prevent many of the infections a positive case would normally cause.
However, a number of factors, such as limits on material and human resources, necessitate economical and efficient use of test resources.

\noindent\textbf{Group testing} aims to improve test quality by testing groups of samples simultaneously. We wish to leverage this framework to design practical and efficient COVID-19 tests with limited testing resources. Group testing can be \textbf{\textit{adaptive}} or \textbf{\textit{non-adaptive}}. In the former, tests can be decided one at a time, taking into account previous test results. In the latter, one can run tests in parallel, but also has to select all tests before seeing any lab results. 

A popular example of a \textbf{\textit{semi-adaptive}} group test is to first split $n$ samples into $g$ groups of (roughly) equal size, pool the samples within the groups and perform $g$ tests on the pooled samples. All samples in negatively tested pools are marked as negative, and all samples in positively tested pools are subsequently tested individually.

\noindent\textbf{Practical Constraints for COVID-19.} Although group testing is a well-studied problem, the recent COVID-19 pandemic introduces specific constraints. In contrast to seroprevalence antibody tests, PCR tests aim to detect {\em active} cases, and only successfully do so during part of the disease course \cite{he2020temporal}). This results in a small {\bf prevalence} (prior probability of population infection; we will assume a {\bf default value} of $10^{-3}$), assuming we screen the general population rather than only symptomatic individuals.
Group testing has recently been validated for COVID-19 PCR tests \cite{Schmidt2020.04.28.20074187,evaluation2020yelin}. It is facilitated by the fact that PCR is an amplification technique that can detect small virus concentrations. Nevertheless, there are limitations on the number of samples $l$ that can be placed in a group (\cite{evaluation2020yelin} considers up to 64), and constraints on the number of times a particular sample can be used (\cite{Schmidt2020.04.28.20074187} uses serial incubation of the same respiratory sample in up to $k=10$ tubes). Besides, there are practical issues: adaptive testing is time consuming and hard to manage. Complex multiplex designs are prone to human error.

Existing research on non-adaptive group testing is generally concerned with identifying at most $k$ positive samples amongst $n$ total samples, which is referred to as non-adaptive {hypergeometric} group testing \cite{hwang1987non}. This assumption yields asymptotic bounds on the number of tests needed to recover the ground truth \cite{knill1998non,indyk2010efficiently,cheraghchi2012graph,chan2014non}. However, these are of limited practical relevance when constructive results on small numbers of samples are required. The specific constraints for COVID-19 force us to revisit the general framework of group testing.

\begin{figure}[tb]
    \centering
    \includegraphics[height=2in]{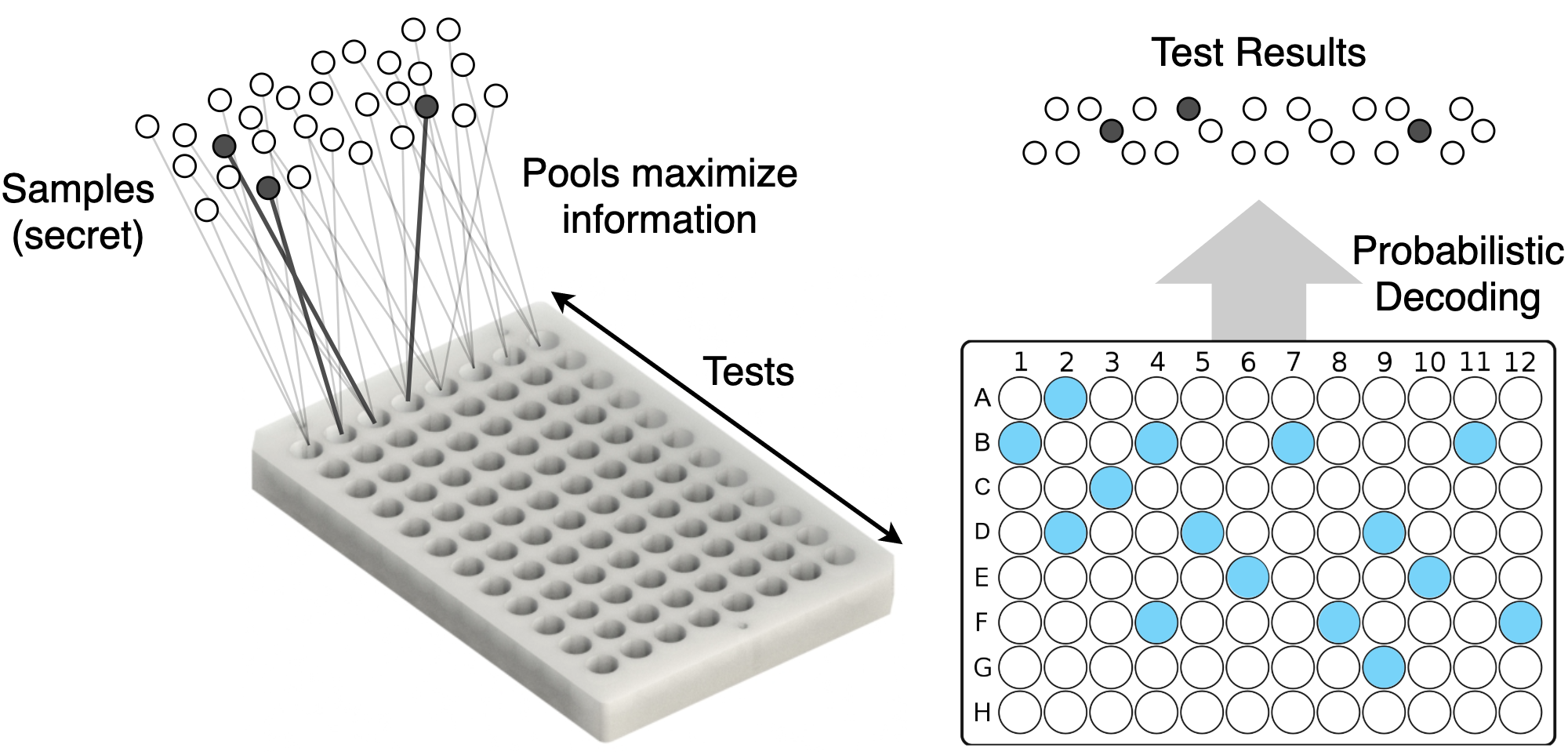}
    \caption{We formulate the group testing problem as a constrained information maximization problem. Samples are grouped into testing pools so that the information gain is maximized while obeying practical constraints (i.e. no more than 64 samples in one group). Here, positive samples are shown in black and positive tests are shown in blue. The tests are decoded with error correcting probabilistic methods.}
    \label{fig:methodology}
\end{figure}

\noindent\textbf{Novel Formulation.} We formulate the problem based on the principle of information gain: given $n$ people and $m$ testing kits, the characteristics of the test and prior probabilities for each person to be sick, we seek to optimize the way the tests are used by combining several samples. For simplicity, samples are assumed to be independent analogous to \cite{mazumdar2016nonadaptive}. However, we focus on implementable tests, unlike~\cite{mazumdar2016nonadaptive} which focuses on asymptotic results that are valid for large $n$. Figure~\ref{fig:methodology} summarizes our approach. 

\noindent\textbf{Optimal Characterization:} By leveraging the framework of adaptive sub-modularity initially developed for sensor covering by \cite{golovin2011adaptive}, we prove near-optimality of a simple greedy-strategy for adaptive testing. 
Despite the simplicity, it turns out that this greedy strategy has exponential running time and becomes infeasible for $n \ge 16$. Fortunately, the near optimally of the greedy-adaptive method points toward a simple and scalable non-adaptive solution leveraging randomization akin to the Bloom Filter structure~\cite{mitzenmacher2017probability}.

\noindent\textbf{Bloom Origami Assays:}\footnote{The term {\em Origami} stems from the idea to use paper folding techniques for test multiplexing, see \cite{origami}.}~\cite{mich2020bloom,broder2020note} recently showed that pooling using random hash functions, similar to a Bloom filter structure, can lead to an efficient and straightforward group testing protocol. We will show that such a protocol, based on random hash functions, is unnecessarily restrictive. 
Bloom filters were designed for streaming data, where there is no choice but to use universal hash functions for pooling. For COVID-19, the computational situation is much simpler. Leveraging our information gain framework, we propose superior but straightforward hashing strategies. 

A bigger problem with Bloom filters is the (necessarily) simple decoder. The decoder trades accuracy for efficiency, as it was designed for internet-scale problems where linear time decoding is prohibitive. For COVID-19, we instead propose a message-passing decoder, similar to Counter Braids \cite{lu2008counter}, which is more accurate. Our proposal of connecting probabilistic graphical model (PGM) inference with Bloom filters could be broadly applicable to situations beyond COVID-19 group testing. Since the graphical model is a bipartite graph for which no narrow junction tree can be found, message passing does not necessarily converge to the global optimum. Therefore, we propose a new method for graphical model inference leveraging probabilistic concentration and meet-in-the-middle (MITM) techniques, which may be of independent interest. Our MITM method is particularly useful for the low prevalence scenario. This paper illustrates the power of algorithmic re-design to target practical constraints. We obtain significant gains even on the relatively well-studied topic of group testing.


\vglue-1mm
\section{Preliminaries}
\vglue-3mm

{Notations} are progressively introduced throughout but are gathered in the appendix, which also contains the proofs. Denote the number of patient\footnote{For simplicity, we will refer to all individuals being tested as {\em patients}.} samples by $n$. As previously mentioned, we consider the group testing task in the particular context of the COVID-19 pandemic. This choice of problem setting naturally introduces new mathematical constraints of a practical nature:

\textbf{Impracticality of Adaptivity.} Adaptive methods require several hours in between each lab result of the adaptive sequence. This inspires us to only consider either non-adaptive methods or semi-adaptive methods with no more than two phases of testing.

\textbf{Low Concentration and Test Accuracy.} Excessive mixing of patient swabs may result in prohibitively low viral concentration with negative consequences for testing. 
A recent study reports that one can safely mix a patient swab up to 10 times \cite{Schmidt2020.04.28.20074187}; another relays that mixing up to 32 patient samples into the same probe yields a false negative rate below 10\% \cite{evaluation2020yelin}. 

There is clearly ambiguity in the limitations of the experimental protocol. For instance, \cite{Smola} validate double-digit numbers of patients per sample for PCR tests. While dilution effects are relevant for such large pools, they can be partly addressed by incubating respiratory swabs multiple times \cite{Schmidt2020.04.28.20074187}. Also note that we are only concerned with the accuracy of the \emph{tests} per se rather than the biological sampling protocol (i.e. whether swabs are taken when viral load is detectable in patients). In what follows we consider group sizes of $n = 100$ as a sensible upper limit. 


\paragraph{Notations and Reminders}
Denote the number of tests to run by $m$. Tests are assumed to be imperfect, with a {\em true positive rate (or sensitivity) $\mathrm{\mathrm{tpr}}$}\footnote{equivalent terms include {\em hit rate}, {\em detection rate} and {\em recall}.} and {\em true negative rate (or specificity) $\mathrm{\mathrm{\mathrm{tnr}}}$}.\footnote{equivalent terms include {\em correct rejection rate} and {\em selectivity}.} 
As simple {\bf default values}, we will use $\mathrm{\mathrm{tpr}}=99\%$ \cite{Padhye2020.04.24.20078949} and $\mathrm{\mathrm{\mathrm{tnr}}} = 90\%$ \cite{evaluation2020yelin}.\footnote{This number is affected by selection bias since it heavily depends on the stage of the disease; it is lower if a person is tested too late \cite{he2020temporal,Padhye2020.04.24.20078949}; our results provide guidance as to how to analyze the samples that were collected rather than the collection timing and protocol itself.}
 
Patient sample $i$ is infected with probability $p_i\in [0,1]$ and we assume statistical independence of infection of patient samples. Denoting by a `1' a positive result (infection), the unknown ground truth is a vector of size $n$ made up of `0's and `1's. This vector describes who is infected and who is not. We call this the \textit{secret}, denoted as $s\in\{0,1\}^n$. A \textit{design of a test} $d\in\{0,1\}^n$ to run in the lab is a subset of patient samples to mix together into the same sample, where $d_i=1$ if patient sample $i$ is mixed into design $d$ and $d_i=0$ otherwise. Note that the outcome of a perfect design $d$ for a given secret $s$ can simply be obtained as $\mathbf{1}_{\langle d,s\rangle > 0}$ where $\langle d,s\rangle:=\sum_{i=1}^n d_i s_i$. That is, a test result is positive if there is at least one patient $i$ for which $d_i=1$ (patient $i$ is included in the sample) and $s_i=1$ (patient $i$ is infected). Figure~\ref{fig:methodology} illustrates the problem setting. 

Recall that the secret $s$ is unknown. However, since we assume that patient sample $i$ is infected with probability $p_i$ and that patient samples are independent, 
we have a \textit{prior} probability distribution over the possible values of $s$.  We hence represent the random value of $s$ as a \textit{random variable (r.v.)}, denoted by $S$, with probability distribution $p_S(s) := \Pr[S=s]$ over $\{0,1\}^n$. Let us now recall the definition of the \textit{entropy} of our random variable,
\begin{equation}
    H(S) = -\sum_{s\in\{0,1\}^n} p_S(s)\log_2 p_S(s),
\end{equation}
The entropy represents \textit{the amount of uncertainty that we have on its outcome}, measured in bits. It is maximized when $S$ follows a uniform distribution, and minimized when $S$ constantly outputs the same value. As we perform tests, we gain additional knowledge about $S$. For instance, if we group all samples into the same pool and have a negative result, then our {\em posterior} probability that all patients are healthy goes up. That is, $p_S((0,\dots,0))$ increases according to Bayes' rule of probability theory. More generally, we may perform a sequence of tests of varying composition, updating our posterior after each test. Our goal will be to select designs of tests so as to minimize entropy, resulting in the least amount of uncertainty about the test outcome for all individuals.



\vglue-1mm
\section{Solving for Small Number of Patients}
\vglue-3mm
Given $n$ people, test characteristics $\mathrm{\mathrm{tpr}}$ \& $\mathrm{\mathrm{\mathrm{tnr}}}$ and a set of prior probabilities of sample infection $(p_i)_{1 \leq i \leq n}$, the best multiset $\mathcal{D}$ of $m$ pool designs is the one maximizing the information gain. The tests are order insensitive, which gives a search space of cardinality ${2^n + m \choose m}$. Evaluating the information gain of every multiset separately takes $\mathcal{O}\left(2^{n+m}\right)$ operations.\footnote{We chose to implement a version with complexity $\mathcal{O}\left(m 2^{n+m}\right)$, but more cache efficient in practice.} Hence, brute-forcing this search space is prohibitive even for small values of $n$ and $m$.

We resort to randomized algorithms to find a good enough solution. Our approach is to use Evolutionary Strategies (ES). We apply a variant of the $(1+\lambda)$ ES with optimal restarts \cite{luby1993optimal} to optimize any objective function over individuals (multisets of tests).

\paragraph{Detailed Description.} We maintain a population of $1$ individual between steps. At every step of the ES, we mutate it in $\lambda \in \mathbb{N}^+$ offsprings.
In the standard $(1+\lambda)$ ES, each offspring is mutated from the population, whereas our offsprings are iteratively mutated, each one being the mutation of the previous. These offsprings are added to the population, and the best element of the population is selected as the next generation of the population.

We initialize our population with the ``zero'' design that doesn't test anyone. Our mutation step is straightforward: flipping one bit $d_i$ of one pool design $d$, both chosen uniformly at random. We also restrict our search space if needed: the number of 1's in a column must be less than the number of times a given swab can be mixed with others, the number of 1's in a line is constrained not to put too many swabs into the same pool. Our iterative mutation scheme allows us to step out of local optima.


After choosing a basis $b$ proportional to $n \times m$ (which is approximately the logarithm of our search space), we apply restarts according to the Luby sequence: $(b,b,2 b,b,b,2 b,4 b,b,b,2 b,b,b,2 b,4 b,8 b,...)$. This sequence of restarts is optimal for Las Vegas algorithms \cite{luby1993optimal}, and our ES can be viewed as such under two conditions: \textit{(i)} that the population never be stuck in a local optimum, which can be achieved in our algorithm using $\lambda = n \times m$ (note that much smaller constant values are used in practice); \textit{(ii)} the second condition is purely conceptual and consists in defining a success as having a score larger than some threshold. The fact that our algorithm does not use this threshold as an input yields the following result, proved in Appendix~\ref{sec:proof-ES}:

\begin{theorem}\label{thm:ES}
Under condition \textit{(i)}, the evolutionary strategy using the Luby sequence for restarts yields a Las Vegas algorithm that restarts optimally \cite{luby1993optimal} to achieve any target score threshold.
\end{theorem}

\vglue-1mm
\section{Motivating Greedy Information Maximization} \label{sec:mutual-info}
\vglue-3mm

Note that since tests are imperfect, for a given pool design $d\in\{0,1\}^n$ and a given secret $s\in\{0,1\}^n$, the Boolean outcome $T(s, d)$ of the test in the lab is not deterministic. 
If tests were perfect, we would have $T(s,d)=\mathbf{1}_{\langle d,s\rangle > 0}$. To allow for imperfect tests, we model $T(s,d)$ as a r.v.\ whose distribution is described by
$\Pr[T(s,d) = 1 \mid \langle d,s\rangle > 0 ] = \mathrm{\mathrm{tpr}}$ and $\Pr[T(s,d) = 0 \mid \langle d,s\rangle = 0 ] = \mathrm{\mathrm{\mathrm{tnr}}}$.\footnote{\label{footnote4}For prior information on whether and how the errors depend on the number of samples mixed into a given pool design (e.g.\ by dilution effects), we can take this into account by letting $\mathrm{\mathrm{tpr}}$ and $\mathrm{\mathrm{\mathrm{tnr}}}$ depend on $|d|=\sum_i d_i$.} Since the secret $s$ is also unknown (and described by the r.v.\ $S$), the outcome $T(S,d)$ has now two sources of randomness: imperfection of tests and unknown secret.\footnote{Laboratory errors in composing the pooled designs $d$ could be modeled by correspondingly describing $d$ by a random variable, or by including these errors into the random variable $T$.} In practice, one will not run one test but multiple tests. We now suppose that $m$ tests of pool designs are run and let their designs be represented as a multiset $\mathcal{D} \in (\{0, 1\}^n)^m$.

This leads us to the following question: given an initial prior probability distribution $p_S$ over the secret, how should we select pool designs to test in the lab? We want to select it such that once we have its outcome, we have as much information as possible about $S$, i.e. the entropy (uncertainty) of $S$ has been minimized. Since we cannot know in advance the outcome of the tests, we have to minimize this quantity \textit{in expectation} over the randomness coming from both the imperfect test and unknown secret. This requires the notion of \textit{conditional entropy}.


\paragraph{Conditional Entropy.} Given pool designs $\mathcal{D}$, we consider two random variables $S$ (secret) and $T:=T(S,\mathcal{D})$ (test results). The conditional entropy of $S$ given $T$ is given by:

\begin{equation}\label{eq:conditional}
H(S | T) = - \hspace{-9mm} \sum_{{s\in\{0,1\}^n, t\in\{0,1\}^m}} \hspace{-7mm} \Pr[S=s, T=t] \cdot  \log_2\left(\frac{\Pr[S=s, T=t]}{\Pr[T=t]}\right) = \mathbb{E}_{t  \sim T(S, \mathcal{D})} \left[H(p_{S | T = t}) \right]
\end{equation}

In this formula, the joint probability $\Pr[S=s, T=t]$ has been computed with the conditional probability formula $\Pr[S=s, T=t] = \Pr[S=s] \Pr[T=t | S=s]$, and the posterior distribution is computed using Bayesian updating, i.e., 
\begin{equation}\label{eq:posterior}
    p_{S | T = t}(s) = \Pr[S=s | T=t] = {\Pr[S=s, T=t]}/{\Pr[T=t]},
\end{equation} where $\Pr[T=t] = \sum_s \Pr[S=s, T=t]$. It represents the amount of information (measured in bits) needed to describe the outcome of $S$, given that the result of $T$ is known. The \textit{mutual information} between $S$ and $T$ can equivalently be defined as 
$I(S,T) := H(S) - H(S|T)$.
It quantifies the amount of information obtained about $S$ by observing $T$.

\paragraph{A well-motivated criterion for test selection.} Since $H(S)$ does not depend on $d$, selecting the pool design $d$ minimizing the conditional entropy of $S$ given the outcome of $\mathcal{D}$ is equivalent to selecting the one maximizing the mutual information between $S$ and $T(S, \mathcal{D})$. We now have a clear criterion for selecting $\mathcal{D}$:
\begin{equation}\label{eq:criterion}
    \mathcal{D}^* \in \arg\max_\mathcal{D} I(S, T(S, \mathcal{D})).
\end{equation}
This criterion selects the pool designs $\mathcal{D}$ whose outcome will maximize our information about $S$.

\paragraph{Expected Confidence.} We report another evaluation metric of interest called the \textit{expected confidence}. It is the mean average precision of the maximum likelihood outcome. The maximum likelihood outcome it defined by:
\begin{equation}
    \mathrm{ML}(t) := \arg\max_s \Pr[S=s | T=t],
\end{equation}
which yields the following definition of Expected Confidence
$\mathrm{Confidence}(S|T) := \Pr[S = \mathrm{ML}(T)]$
\begin{align}
    \Pr[S = \mathrm{ML}(T)]
    = \sum_{t\in\{0,1\}^m} \Pr[T=t, S=\mathrm{ML}(t)]
    =\mathbb{E}_{t  \sim T(S, \mathcal{D})} \left[\max_s p_{S | T=t}(s)\right]
\end{align}
$ML$ is of particular practical interest: given test results $t$, a physician wants to make a prediction. In this case, it makes sense to use the maximum likelihood predictor.
The interpretation of $\mathrm{Confidence}$ is straightforward: it is the probability that the prediction is true (across all possible secrets).

\paragraph{Updating the priors.} Both scoring functions described above compute the expectation relative to the test results of a score on the posterior distribution $p_{S | T=t}(s)$. After observing the test results, we are able to replace the prior distribution $p_S$ by the posterior. By the rules of Bayesian computation, this update operation is commutative, i.e., the order in which designs $d_1$ and $d_2$ are tested does not matter, and compositional in the sense that we can test $\{d_1, d_2\}$ simultaneously with the same results. 
Thus, we can decompose those steps and make different choices as we run tests (see the adaptive method below).

Although searching the space of all possible adaptive strategies would yield a prohibitive complexity of $\Omega(2^{2^m})$, it turns out that a simple adaptive strategy can yield provably near-optimal results. We describe an adaptive scheme in Algorithm~\ref{alg:greedy} which greedily optimizes the criterion defined in Eq.~(\ref{eq:criterion}).

\begin{algorithm}[H]
\SetAlgoLined
\textbf{Input:} Numbers $n$ \& $m$, test characteristics $\mathrm{\mathrm{tpr}}$ \& $\mathrm{\mathrm{tnr}}$, priors $p_i$ for $i\in\{1,...,n\}$\;
\textbf{Output:} The sequence of tests to adaptively run in the lab\;
\textbf{Initialization:} Set $k:= m$ and set prior $p_S$ using the $p_i$'s\;
 \While{$k>0$}{
 For each pool design $d$ in $\{0,1\}^n$, compute $I(S,T(S, d))$\;
 Select any 
 $d^*\in \arg\max_d I(S,T(S, d))$\;
 Observe result $T(S, d^*)$ of design $d^*$ in the lab\;
 Update $p_S$ accordingly (see Eq.~(\ref{eq:posterior})) 
 to the realization of $d^*$ in the lab \;
 Decrease the number of remaining tests $k$ by $1$\;
 }
 \caption{(Greedy-Adaptive)}\label{alg:greedy}
\end{algorithm}
Leveraging the framework of adaptive sub-modularity \cite{golovin2011adaptive}, and assuming that the criterion defined by Eq.~(\ref{eq:criterion}) is adaptive sub-modular\footnote{Empirical validation in Appendix~\ref{sec:appendix-submodular-assumption}.}, Algorithm~\ref{alg:greedy} has the guarantee below. 
\begin{theorem}\label{thm:greedy} Denote by `$\mathrm{Algo}$' an adaptive strategy. Let $I(\mathrm{Algo})$ be the expected mutual information obtained at the end of all $m$ tests by running $\mathrm{Algo}$, the expectation being taken over all $2^m$ outcomes of lab results. Denote by `Optimal' the best (unknown) adaptive strategy. If we run Algorithm~\ref{alg:greedy} for $m_1$ tests and Optimal for $m_2$ tests, we have:
\begin{equation}
    I(\mathrm{Algorithm~\ref{alg:greedy}}) \geq \left(1-e^{-\frac{m_1}{\alpha m_2}}\right)I(\mathrm{Optimal}),
\end{equation}
where $\alpha$ is defined as follows: assume that our priors $p_i$ are wrong, in the sense that there exist constants $c,d$ with $c p_i\leq p_i'\leq d p_i$ for $i\in\{1,...,n\}$, with $c\leq 1$ and $d\geq 1$, where $p_i'$ denotes the true prior: we set $\alpha:=d/c$. 
\end{theorem}
\paragraph{Remarks.} Accordingly, Algorithm~\ref{alg:greedy} is \textit{(i)} robust to wrong priors and \textit{(ii)} near-optimal in the sense that the ratio of its performance with that of the optimal strategy goes to $1$ exponentially in the ratio of the numbers of tests run in each algorithm. For $\alpha=1$ and $m_1=m_2$, this yields $1-e^{-1}\simeq 0.63$.

\vglue-1mm
\section{Testing at Scale with Bloom Filters} \label{sec:bloom-filters}
\vglue-3mm


Our previous methods are effective, but they are prohibitively expensive for $n > 30$ patients. To address this, we present a randomized approach to selecting $\mathcal{D}$ by grouping patients into pools using Bloom filters \cite{Bloom}.

\begin{wrapfigure}[13]{r}{1.5in}
  \vspace{-0.8cm}
  \begin{center}
    \includegraphics[height=1.5in]{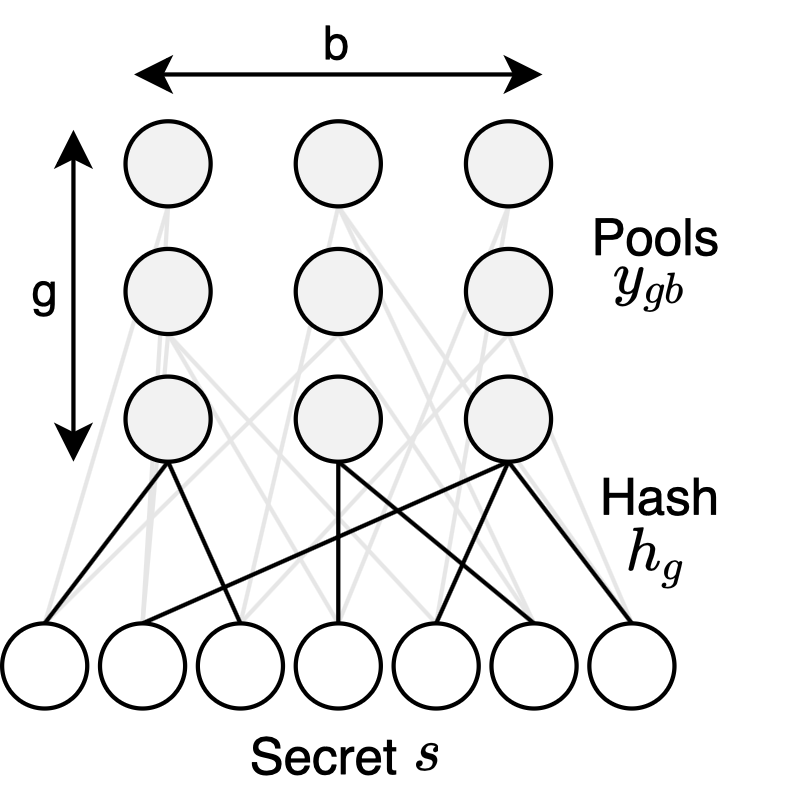}
  \end{center}
  \vspace{-0.15in}
  \caption{Test design. $n$ people are shuffled $g$ times and divided into $b$ bins. \vspace{-0.1cm}}
  \label{fig:pgm}
\end{wrapfigure}

Randomized test pooling may be attractive to practitioners because it is straightforward to understand and implement in the laboratory. The simplest method partitions $n$ patients into random groups of equal size. Patients are either re-tested or reported positive if their group tests positive ({\bf Single Pooling}). In~\cite{broder2020note}, the authors propose an extension to this idea that inserts patients into two sets of pools, named {\bf double pooling}, which offers impressive advantages at the same cost. We present a generalization of this idea that uses an array of Bloom filters to improve the error characteristics of the test. While Bloom filters have been considered for the low-prevalence COVID-19 testing problem ~\cite{mich2020bloom,gitCovidBloom}, current methods are based on a simple randomized encoding and decoding process that was designed for internet-scale applications where even linear time was prohibitive and where the keys are not known beforehand. This sacrifices accuracy. We now design an improved algorithm.



\paragraph{Encoding.}
Bloom filters use universal random hash functions for load balancing because the streaming algorithm setting does not allow us to control the number of items in each group.
Here, we can improve the filter with perfect load balancing.
We divide the $m$ tests into $g$ groups of $b$ pools. In each group, we assign the $n$ patient samples to the $b$ pools so that each pool contains $n/b$ patients.\footnote{or $\lfloor n/b\rfloor$ and $\lfloor n/b\rfloor + 1$ patients if $n$ is not a multiple of $b$.}. This procedure constrains the multiset $\mathcal{D}$ of possible test designs. With uniform prior probabilities, we implement a perfectly load balanced hash by assigning each patient a number based on a permutation $\pi_j$ of the integers $\{1,...n\}$ Thus patient $i$ is assigned to pool $h_j(i):=\pi_j(i)\,\mathrm{mod}\,b$ in group $j$. 

For non-uniform priors, we can resort to a variable load hash to balance total weights into pools. Due to the concavity of the entropy, the information gain is maximized if all pools have the same probability of testing positive. This is maximized for $1/2$, the mode of the binary entropy. 

\paragraph{Load balancing implies Information Gain:} Load balancing, as exhibited by our encoding, maximizes the information gain for a practical subset of constrained Bloom filter group test problems. Theorem~\ref{thm:bloom_info_gain} motivates Bloom filters in the context of our information theoretic framework. With a constraint on the number of samples in each pool, our load balancing hash allocation is the optimal pooling strategy provided that $\mathrm{Pr}[t_b = 1]$ is sufficiently small ($\sim 20$\%). We defer a detailed discussion to the appendix.

\begin{theorem}
Assuming independent priors, the information gain $I(S,T)$ of the tests $\{t_1, ... t_b\}$, in a single Bloom filter row is maximized by having all the positive pool probabilities $\mathrm{Pr}[t_b = 0] = \prod_{i\in \text{pool }b}\mathrm{Pr}[s_i = 0]$ equal to a constant that depends only on tpr and tnr.
\label{thm:bloom_info_gain}
\end{theorem}



\paragraph{Decoding for Perfect Tests.} 
Assuming perfect tests, one can easily decode the pooled test results $t\in\{0,1\}^{b\times g}$ because all patients in negative pools are healthy. We can then identify positive (and ambiguous) samples by eliminating healthy samples from positive pools, as described in the appendix. In the case where $g = 1$ and $g = 2$, we have the widely-used single pooling method and the recently-proposed double pooling method~\cite{broder2020note}. Assuming there are no false negative pool results, one can use the decoder to identify all positive samples and derive optimal dimensions $b\times g$ that minimize the number of tests, as shown in the below theorem:
\begin{theorem}
\label{thm:bloom_dimensions}
Given $m$ perfect tests, $n$ patients and a uniform prior (prevalence) $\rho$, the decoder correctly identifies all positive samples and mislabels any negative sample with probability 
$\mathrm{P}[\hat{s}_i = 1 | s_i = 0] \leq \left(1 - e^{-\rho\frac{n}{b}}\right)^g$.
The bound is minimized for $g = \frac{m}{n\rho}\log 2$ and $b = m / g$. 

\end{theorem}
The analysis borrows tools from regular Bloom filters and the results shown in \cite{mitzenmacher2017probability}. Note that the problem with no test error and $1/2$ prevalence is a \#P-complete restriction of \#SAT, called monotone CNF \cite{vaismanmodel}.
Realistic tests with nontrivial $\mathrm{fnr}$ and $\mathrm{fpr}$ are technically more interesting. A natural idea is an algorithm dating back to \cite{jaakkola1999variational} when decoding diseases from the QMR database.

\begin{figure}[tb]
    \centering
    \includegraphics[height=2in]{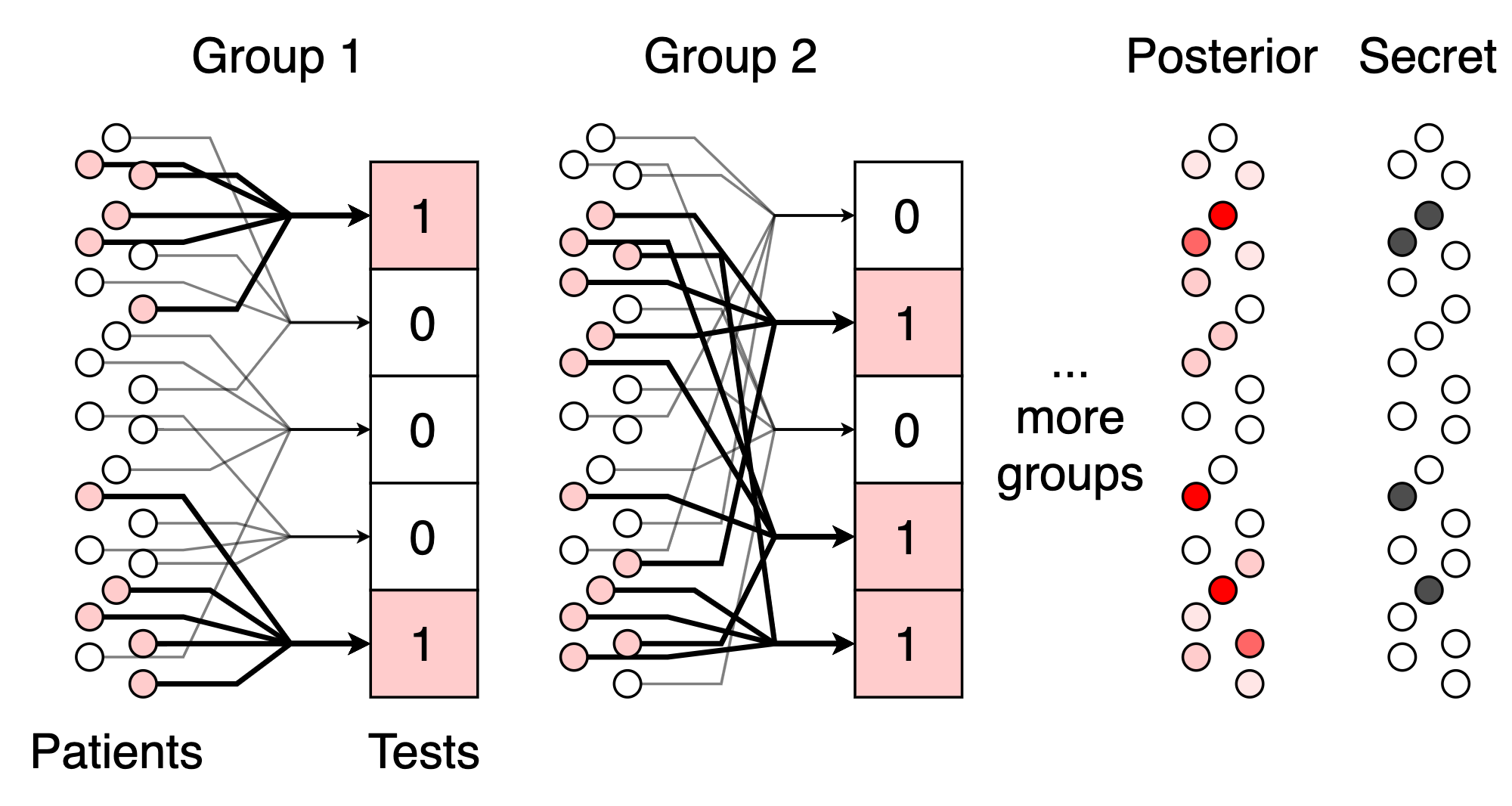}
    \caption{Intuition behind probabilistic decoding. In each group, we suspect that patients in positive pools are positive. If a patient falls within multiple positive pools, the likelihood that their test status is positive increases. Even if a false positive or negative occurs, we may still report the correct diagnosis thanks to information from other groups. This process is known as ``error correction'' and can be implemented with message passing or our MITM algorithm. }
    \label{fig:probabilistic_intuition}
\end{figure}

\paragraph{Decoding via Message Passing.} 

\begin{wrapfigure}{r}{0.45\textwidth}
  \vspace{-0.4cm}
  \includegraphics[width=0.45\textwidth]{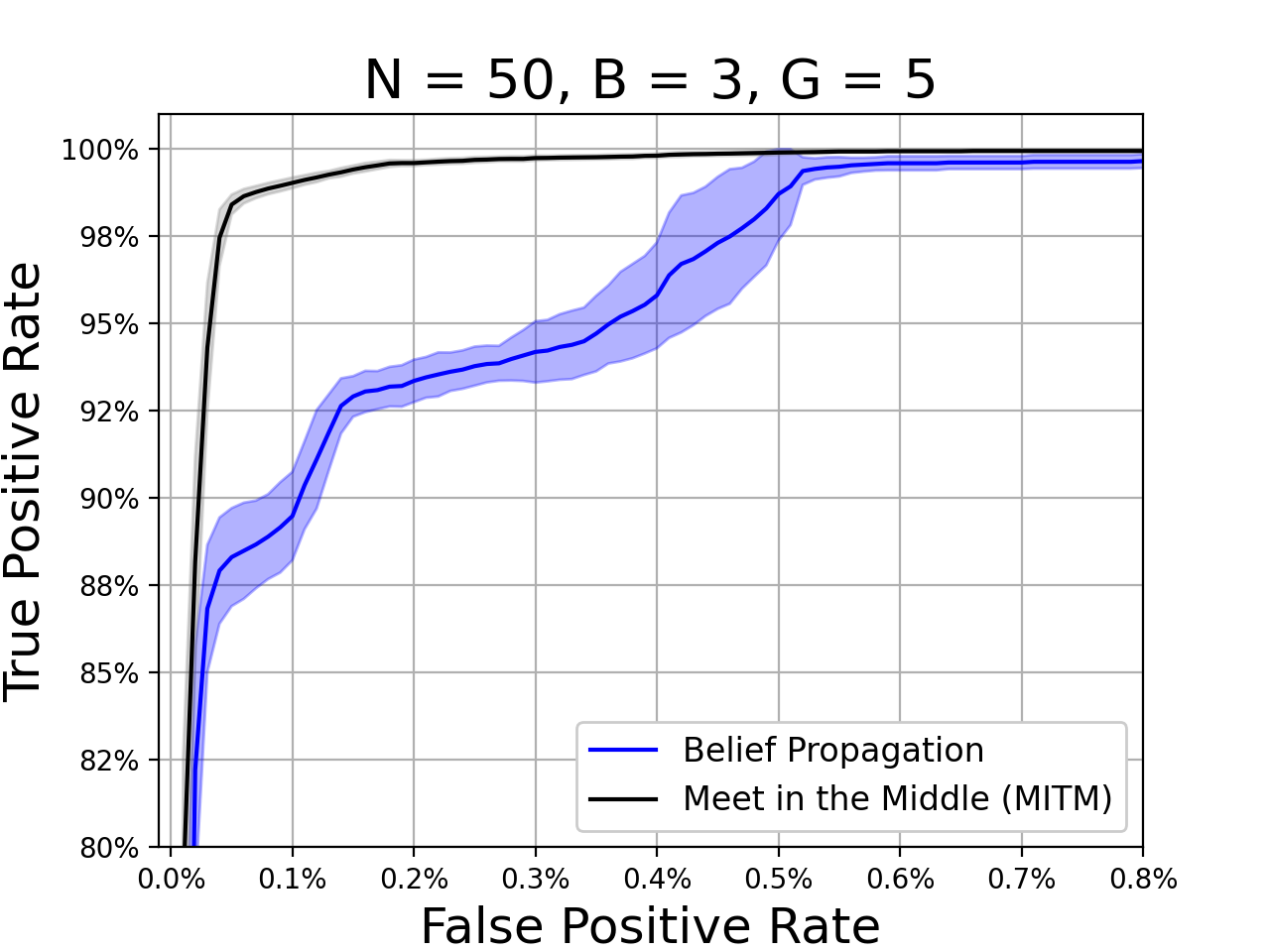}
  \vspace{-0.4cm}
  \caption{Classification using posteriors from different inference methods. }
  \vspace{-0.4cm}
  \label{fig:compare_pgm_approx}
\end{wrapfigure} 

Indeed, false negative rates are often as high as 10\%. The decoder fails for imperfect tests because even negative pools might contain positive samples. A small number of healthy pools might even test positive for some protocols (e.g.\ due to spurious contamination). 


When viewed as a probabilistic graphical model we can interpret  $t_{gb}$ as a corrupted version of the true state $y_{gb}$. It is our goal to infer the secret $s$ that produced $t_{gb}$. Belief propagation is a common technique to estimate the posterior distribution $p_{S|T = t}$ for a graphical model. Since our graphical model cannot be rewritten as a junction tree with narrow tree width there are no efficient exact algorithms. Instead, we resort to loopy belief propagation \cite{koller2009probabilistic}. 

While inexact (loopy-BP isn't guaranteed to converge to the minimum) the resulting solution can classify samples as positive or negative with reasonable performance. While the degree of each pool node is so high that the clique potential would naively involve an intractable number of states, the clique potentials have a simple form that permits an efficient implementation (detail in the appendix). 

\begin{figure}[tb]
    \centering
    \includegraphics[height=2in]{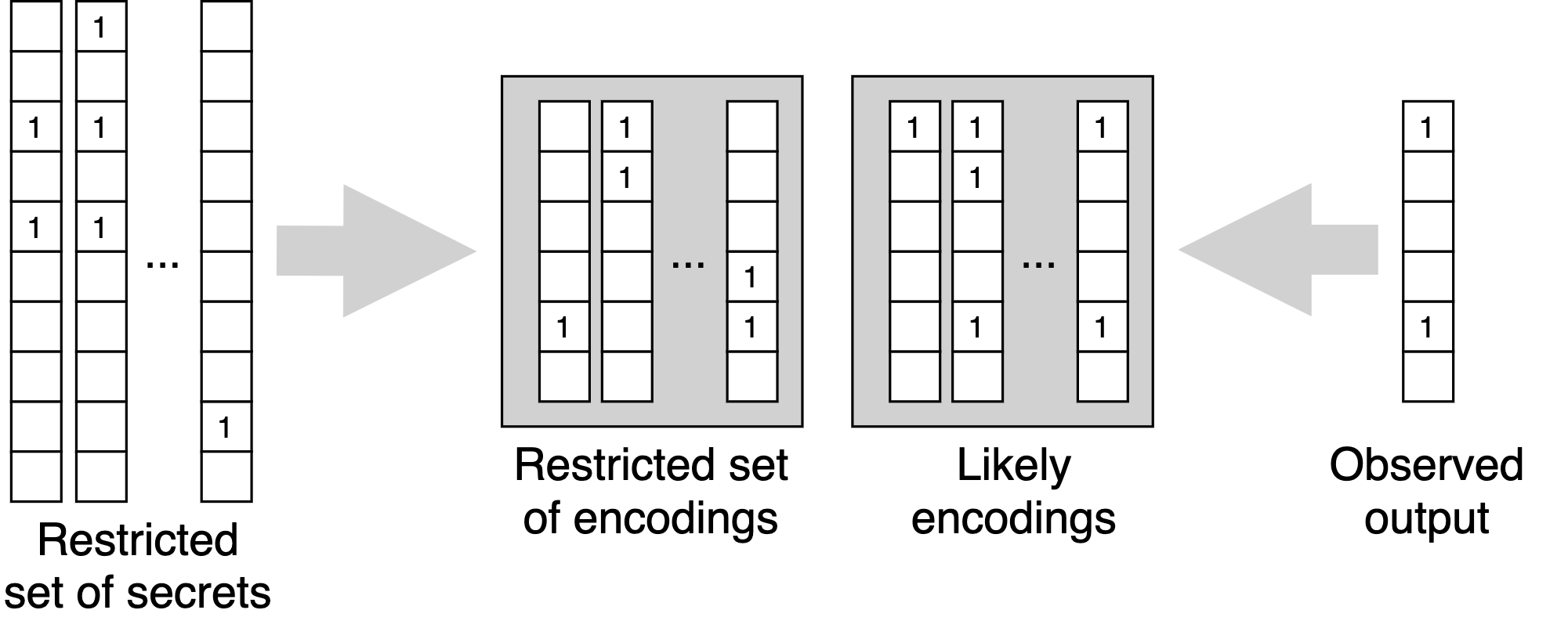}
    \caption{Intuition behind the MITM approach. If the prevalence is low, then we do not need to consider inputs with many positives. This restricts the set of possible secrets and the set of ways we can encode those secrets. The figure shows the inputs for at most 3 positives. Given a (potentially corrupted) output, there are only a small set of true encodings that could have produced that output - it is highly unlikely that every test had a false result. The two conditions ``meet in the middle'' to produce a small set of states. Our MITM algorithm efficiently approximates the posterior probabilities by summing over this restricted state space.}
    \label{fig:mitm_intuition}
\end{figure}

\paragraph{Decoding for Imperfect Tests: Meet-in-the-Middle (MITM).} The structure of the problem also enables an efficient approximation to the exact solution in the (realistic) setting where the tests are fairly accurate and the disease prevalence is low. Low prevalence implies that there are relatively few ``likely secrets'' $s\in\{0,1\}^n$, because most $s_i$ are 0 with high probability. Thus, we only need to consider secrets with a small number of positive patients. 

Since the secrets concentrate in a small subset of $\{0,1\}^n$, we expect to see relatively few Bloom encodings $y \in \{0,1\}^t$ for low-prevalence problems. Furthermore, the output space is likely to be corrupted in relatively few ways. The true state $y_{gb}$ is likely to be the same as the observed output $t_{gb}$, so we only need to consider states that are similar to the observed output. By restricting our attention to ``likely secrets'' and ``likely outputs'', we can reduce the $\mathcal{O}(2^{n})$ complexity of the naive brute-force algorithm. This process constitutes a ``meet in the middle'' approach where we only need to consider a small number of Bloom encodings for inference (Figure~\ref{fig:mitm_intuition}). We show detailed pseudo-code in Algorithm~\ref{alg:mitm-decoder}, and prove Theorem~\ref{thm:mitm} in Appendix~\ref{sec:proof-mitm}.




\begin{theorem}
\label{thm:mitm}
Let $\varepsilon > 0$ and consider the smallest $k$ such that $f(k):=\sum_{j=k}^n {n \choose j} p^j (1-p)^{n-j} < \varepsilon$. Define $A(\varepsilon):=\sum_{i=0}^{k-1}{n \choose i}$, and $C(\varepsilon)$ the number of different encodings of secrets with less than $k$ infected people. Obviously\footnote{Because the code space is the image of the secret space w.r.t. the encoding function.}, $C(\varepsilon) < A(\varepsilon)$ and in practice $C(\varepsilon) \ll A(\varepsilon)$. For any test result $t \in\{0,1\}^m$, define $P:=\sum_i t_i$ and $N:=m-P$. Let $B(\varepsilon) := \sum_{prob[FP][FN]>\varepsilon} {P \choose FP}{N \choose FN}$, where
\begin{equation}
    prob[FP][FN]:= (1-\mathrm{tnr})^{FP}\mathrm{\mathrm{tpr}}^{P-FP}(1-\mathrm{\mathrm{tpr}})^{FN}\mathrm{\mathrm{tnr}}^{N-FN}.
\end{equation}

Then there exists an algorithm with preprocessing time $\mathcal{O}((n+m) A(\varepsilon))$, space complexity $\mathcal{O}((n+m) C(\varepsilon))$ and query time $\mathcal{O}((n+m)  \min(B(\varepsilon), C(\varepsilon)))$ that estimates $P[s_i | t]$ as a fraction $\tilde P[s_i \land t] / \tilde P[t]$ with an error less than $4\varepsilon/P[t]\leq 4\varepsilon/ \tilde P[t]$, where the $\tilde P$ values are our estimation of $P$.
\end{theorem}

\paragraph{Illustrative values for MITM decoding.} Using Stirling's formula $n!\sim \sqrt{2\pi n}(n/e)^n$, one can easily show that for a fraction $x\in [0,1]$, we have $f(xn)=o((2p^x)^n)$ when $n\to +\infty$. If $k:=xn$ is such that $2p^x < 1$, i.e. $x>\log(2)/\log(1/p)$, then $f(k)$ will be exponentially small w.r.t. $n$. With our default $p=0.1\%$ we only need to consider secrets with a fraction smaller than $x^*=\log_2(2)/\log_2(1/10^{-3})\approx 10.03\%$ of infected people to yield negligible error. For $n=60$, choosing $x=13\%$ reduces\footnote{We actually observe much tighter bounds in practice.} the search space of secrets from $2^{60}\approx 10^{18}$ to $\sum_{i=0}^{\lceil 60*0.13\rceil -1}{60 \choose i} < 6\cdot 10^7$ with an error $\varepsilon < (2p^{0.13})^{60} < 5\cdot 10^{-6}.$

\vglue-1mm
\section{Numerical Experiments}
\vglue-3mm

\begin{figure}[tb]
    \centering
    \includegraphics[width=0.54\textwidth]{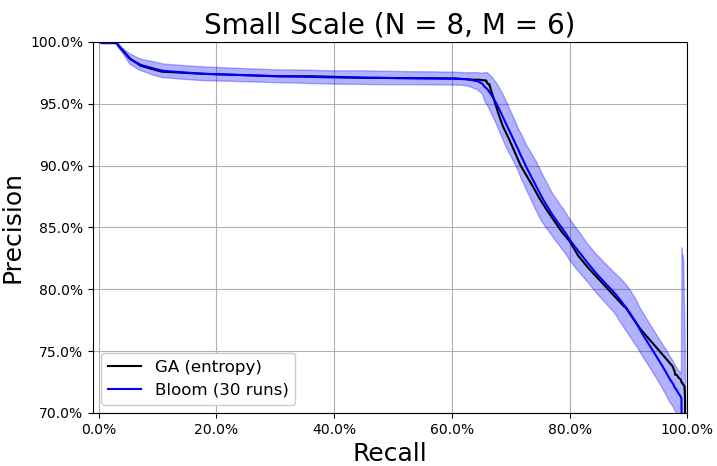}
    \hfill
    \includegraphics[width=0.45\textwidth]{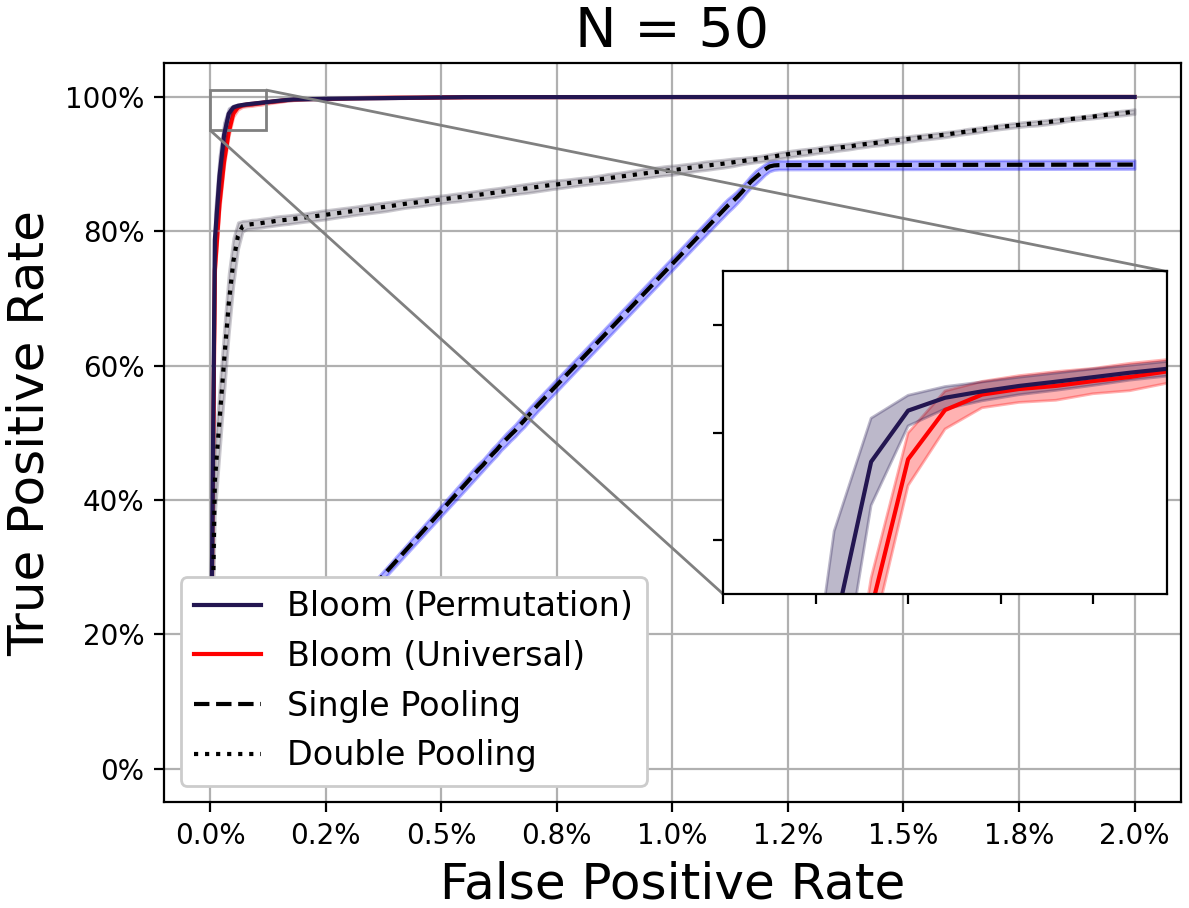}
    \caption{Comparison of group testing designs. We compare brute-force optimization by a genetic algorithm (GA) to our randomized Bloom design on small-scale experiments (left) and our Bloom design against baselines for a larger problem (right). All these experiments used MITM decoding. }
    \label{fig:roc_medium_scale}
\end{figure}


We ran simulations to compare test designs for a large variety of group testing parameters ($n$, tpr/tnr, $b\times g$, prevalence) in the appendix. In this section we present results for a practical scenario where $\mathrm{\mathrm{tnr}} = 0.9$, $\mathrm{\mathrm{tpr}} = 0.99$, and 0.1\% prevalence. In Figure~\ref{fig:roc_medium_scale}, we compare the entropy-minimizing solution found by genetic algorithms with several Bloom filter designs. The Bloom filter performance closely resembles the optimal solution, albeit with higher variance. This validates our claim that the load balancing permutation hash implies a good information gain. 
We also apply our graphical model framework to $3\times5$ arrays of Bloom filters, single pooling and double pooling designs. We use the MITM technique to compute posteriors for all designs and we compare performance. While MITM provides the best results, computational constraints may demand belief propagation for situations where there are many positive group tests. In the high-prevalence scenario, belief propagation will still provide sufficient error correction for good diagnostic results (Figure~\ref{fig:compare_pgm_approx}). The vanilla bloom decoding (single and double pooling) is unnecessarily inaccurate, clearly implying the need for specific tailored algorithms.







\vglue-1mm
\section{Conclusion \& Future Work}
\vglue-3mm
We have presented a framework for group testing taking into account specifics of the current COVID-19 pandemic. It applies methods of probability and information theory to construct and decode multiplex codes spanning the relevant range of group sizes, establishing an interesting connection to Bloom filters and graphical models inference along the way. Our empirical results, more of which are included in the appendix, show that our methods lead to better codes than randomized pooling and popular approaches such as single pooling and double pooling.

Furthermore, we provide an approximate inference algorithm through Theorem~\ref{thm:mitm} that outperforms the message passing approach for realistic parameter values by pruning the exponential search space. We also prove compute-time bounds on its error, highly useful in practice because they are strict.

We believe that the test multiplexing problem is an ideal opportunity for our community to make a contribution towards addressing the current global crisis. By firmly rooting this problem in learning and inference methods, we provide fertile ground for further development. As more information about test characteristics becomes available, we could take into account dependencies of $\mathrm{tpr}$, $\mathrm{\mathrm{tnr}}$ on pool size. The framework could be adapted to different objective functions, or linked to decision theory using suitable risk functionals, e.g., taking into account the downstream risk of misdiagnosing an individual with particular characteristics (comorbidities, probability of spreading the disease, etc.). It can be combined with the output of other methods providing individualized estimated of infection probabilities, to optimize pool allocation for non-uniform priors/prevalence. Statistical dependencies (e.g., for family members) could be taken into account. Finally, similar methods also permit addressing the problem of prevalence estimation. Further details as well as some concrete design recommendations derived from our methods are available in the appendix.

\section*{Acknowledgments}
Gary B\'ecigneul is funded by the Max Planck ETH Center for Learning Systems. Benjamin Coleman and Anshumali Shrivastava are supported by NSF- 1652131, Nsf-BigData 1838177, AFOSR-YIPFA9550-18- 1-0152, Amazon Research Award, and ONR BRC grant for Randomized Numerical Linear Algebra.

\clearpage
\section{Broader Impact}
The motivation for this work was to help address the worldwide shortage of testing capacity for Cov-SARS-2. Testing plays a major role in breaking infection chains, monitoring the pandemic, and informing public policy. Countries successful at containing Covid-19 tend to be those that test a lot.\footnote{\url{https://ourworldindata.org/coronavirus-testing}}

On an individual level, availability of tests allows early and targeted care for high-risk patients. While treatment options are limited, it is believed that antiviral drugs are most effective if administered early on, since medical complications in later stages of the disease are substantially driven by inflammatory processes, rather than by the virus itself \cite{inflammatory}. 

Finally, large-scale testing as enabled by pooling and multiplexing strategies may be a crucial component for opening up our societies and economies. People want to visit their family members in nursing homes, send their children to school, and the economy needs to function in order to secure supply chains and allow people to earn their livelihoods.\footnote{\url{http://www.oecd.org/coronavirus/policy-responses/testing-for-covid-19-a-way-to-lift-confinement-restrictions-89756248/}}

However, the present work also poses some ethical challenges, of which we would like to list the below.

The first family concerns the accuracy of the tests. Indeed, when the number of tests and patients are equal, it is natural to compare the $\mathrm{tpr}$/$\mathrm{\mathrm{tnr}}$ of the individual test to the $\mathrm{tpr}$/$\mathrm{\mathrm{tnr}}$ of the individual results in our grouped test framework (obtained by marginalizing the posterior distribution). In some situations with unbalanced priors, the marginal $\mathrm{tpr}$/$\mathrm{\mathrm{tnr}}$ of some people in the group could be lower than the test $\mathrm{tpr}$/$\mathrm{\mathrm{tnr}}$, even if the test will be more successful overall. However, reporting the marginal individual results gives doctors a tool to decide whether further testing should be needed; hence we cannot rule out that individuals might be worse off by being tested in a group. We furthermore show in the appendix that some designs are more fair than others, in that the individual performances are more equally distributed. 

The second family of concerns, directly resulting from the first, is the responsibility of the doctor when assigning the people to batches and giving them prior probabilities (using another model). The assignment of people in batches should be dealt with in a future extension of our framework, while the sensitivity of our protocols to priors should be studied in more depth. The adaptive framework may be more robust with respect to the choice of priors than the non-adaptive one.

Finally, the possibility of truly large scale testing may allow countries with sufficient financial resources to perform daily testing of large populations, with significant advantages for economic activity. This, in turn, could exacerbate economic imbalances.

\bibliographystyle{plain}
\bibliography{biblio}

\include{appendix}
\end{document}

%% file: appendix.tex
\newpage
\appendix

\section{List of All Notations}\label{sec:notations}
We use upper case letters exclusively for random variables (r.v.), except for mutual information $I$ and entropy $H$.
\begin{itemize}
    \item $n$: number of patient samples;
    \item $m$: number of tests to run in the lab;
    \item $g$: number of groups; $b$: number of pools; $m=g\cdot b$;
    \item $s\in\{0,1\}^n$: the \textit{secret} to unveil, with $s_i=1$ if and only if patient sample $i$ is positive (infected);
    \item $S$: r.v. over possible values of $s$ whose law describes the current information we have about $s$;
    \item $d\in\{0,1\}^n$: a pool design, with $d_i=1$ if and only if patient sample $i$ belongs to pool design $d$;
    \item $\mathcal{D}\in(\{0,1\}^n)^m$: random multiset describing the pool designs output by the strategy;
    \item $t\in\{0,1\}^m$: lab result of a list of $m$ tests;
    \item $T$: r.v. over possible values of $t$ describing lab results;
    \item $\mathrm{tpr}$: true positive rate, sensitivity, hit rate, detection rate, recall;
    \item $\mathrm{\mathrm{tnr}}$: true negative rate, specificity, correct rejection rate, selectivity;
    \item $p_i\in [0,1]$: prior probability of infection of patient sample $i$;
    \item $\mathrm{Pr}[A]$: probability of event $A$ to happen;
    \item $p_S(s)\in[0,1]$: probability of secret $s\in\{0,1\}^n$ to be the correct one, according to the law $p_S$ of r.v. $S$.
\end{itemize}

\section{Future Work \& Additional Considerations}

\subsection{Fairness Considerations}

\begin{figure}[h]
    \centering
    \subfigure[Bloom: $(b,g)=(2,3)$]{\includegraphics[scale=0.38]{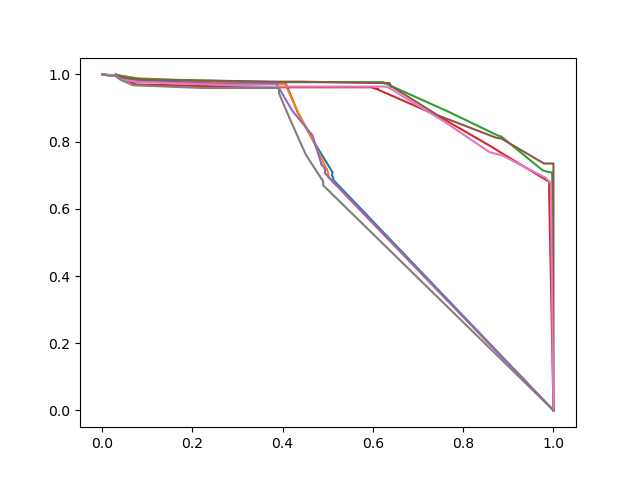}}
    \subfigure[Entropy: $(m,k)=(6,3)$]{\includegraphics[scale=0.38]{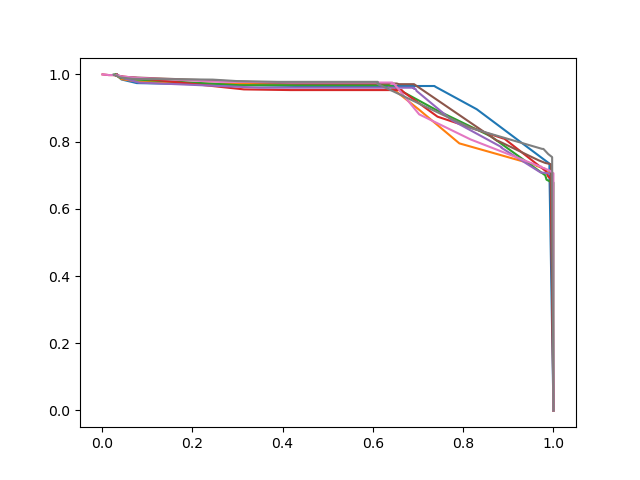}}
     \subfigure[Bloom: $(b,g)=(3,2)$]{\includegraphics[scale=0.38]{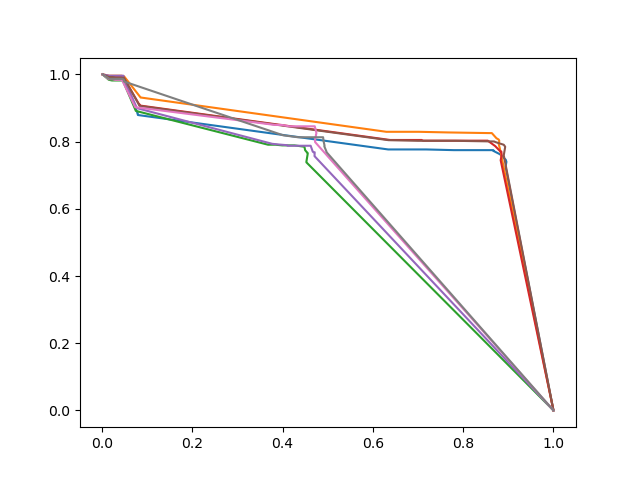}}
    \subfigure[Entropy: $(m,k)=(6,2)$]{\includegraphics[scale=0.38]{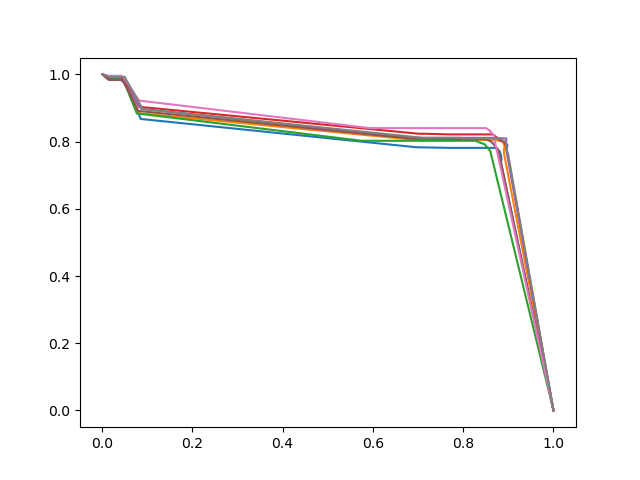}}
    \caption{Precision-Recall curves for different patients, for a set of $n=8$ patients, a prevalence $p=10^{-3}$ and a total of $m=6=b\times g$ tests. Each plot depicts $8$ curves: one per patient. Recall that $b$ denotes the number of bins, $g$ the number of groups, and $k$ the maximum number of times one patient swab can be tested. Hence one should compare (a) with (b), and (c) with (d). ``Bloom'' denotes the use of the Bloom encoding described in Section~\ref{sec:bloom-filters} while ``Entropy'' denotes the use of the Conditional Entropy / Mutual Information encoder described in Section~\ref{sec:mutual-info}. In both comparisons, we observe that the Entropy encoder yields more similar PR curves across patients, compared to the Bloom encoder.}
    \label{fig:fairness}
\end{figure}

Figure~\ref{fig:fairness} illustrates different Precision-Recall curves for different patients, across different methods/parameters. In particular, it shows that the Bloom encoder gives more uneven estimation performances across patients, compared to the Entropy encoder.

\subsection{Others}


\paragraph{Different objective functions.} We have used the number of tests and samples as given, and then optimized a conditional entropy. However, from a practical point of view, other quantities are relevant and may need to be included in the objective, \textit{e.g.} the expectation (over a population) of the waiting time before an individual is ``cleared'' as negative (and can then go to work, visit a nursing home, or perform other actions which may require a confirmation of non-infectiousness). 

\paragraph{Semi-adaptive tests.} Instead of performing $m$ consecutive tests, one could do them in $k$ batches of respective sizes $m_1,...,m_k$ satisfying $m_1+...+m_k=m$.  Adaptivity over the sequence of length $k$ could be handled greedily as in Algorithm~\ref{alg:greedy}, except that instead of selecting a single pool design $d^*$, we would select $m_i$ designs at the $i^{th}$ step. We named this semi-adaptive algorithm the \textit{k-greedy} strategy. 

\paragraph{Further practical considerations.} A good practical strategy could be to perform one round of pooled tests to disjoint groups every morning as individuals arrive at work, being evaluated during work hours. Those who are in a positive group (adaptively) get assigned to a second pool design tested later, which can consist of a non-adaptive combination of multiple designs, tested over night. They receive the result in the morning before they go to work, and if individually positive, they enter quarantine. 
If the test is so sensitive that it detects infections even before individuals become contagious (which may be the case for PCR tests), such a strategy could avoid most infections at work.

\paragraph{Dependencies between $\mathrm{tpr}$, $\mathrm{\mathrm{tnr}}$ and pool size.} The reliability of tests may vary with pool size. In our notation, the outcome of the tests is a random variable that need not only depend on whether one person is sick ($\mathbf{1}_{\langle d,s\rangle > 0}$) but it may also depend on the number of tested people $|d|$ and the number of sick people $\langle d,s\rangle$ (cf.\ Footnote~\ref{footnote4}); it could even assign different values of $\mathrm{tpr}$ and $\mathrm{\mathrm{tnr}}$ to different people. 
The $\mathrm{tpr}$ may in practice be an increasing function of the proportion of sick people ${\langle d,s\rangle} / {|d|}$.

\paragraph{Estimating prior infection probabilities.} Currently, we start with a factorized prior of infection that not only assumes independence between the tested patients but is also oblivious to the individual characteristics. We could, however, build a simple ML system that estimates the prior probabilities based on a set of features such as: job, number of people living in the same household, number of children, location of home, movement or contact data, etc.\footnote{Subject to privacy considerations.} Those prior probabilities can then be readily used by our approach to optimize the pool designs, and the ML system can gradually be improved as we gather more test results.

\paragraph{Prevalence estimation.}
Similar methods can be applied to the question of estimating prevalence. Note that this is an easier problem in the sense that we need not necessarily estimate which individuals are positive, but only how many.

\section{Proofs}
\subsection{Theorem~\ref{thm:ES}}\label{sec:proof-ES}

\paragraph{Statement:} Under the condition that the population never be stuck in a local optimum, the evolutionary strategy using the Luby sequence $(b,b,2 b,b,b,2 b,4 b,b,b,2 b,b,b,2 b,4 b,8 b,...)$ for restarts yields a Las Vegas algorithm that restarts optimally \cite{luby1993optimal} to achieve any target score threshold.
\begin{proof}
Let us remind the main result we use on optimal restarts \cite{luby1993optimal}: the simple Luby sequence of times of restart given by $(b,b,2 b,b,b,2 b,4 b,b,b,2 b,b,b,2 b,4 b,8 b,...)$ is optimal (up to a log factor) for Las Vegas algorithms (i.e. randomized algorithms that always provide the correct answer when they stop, but may have non-deterministic running time). Our theorem is a direct consequence of conceptually casting our problem as a Las Vegas algorithm: indeed, we seek to optimize a fitness function $f$. For a given threshold $A>0$, we can replace the maximization of $f$ by the condition $f>A$. Applying the result of \cite{luby1993optimal} for an exhaustive family of thresholds yields the desired result.
\end{proof}

\subsection{Theorem~\ref{thm:greedy}}\label{sec:proof-greedy}

We wish to invoke Theorem~1 of \cite{golovin2011adaptive}. In order to do so, we need to prove that the conditional entropy which we introduced in Eq.~(\ref{eq:conditional}) is \textit{adaptive monotone}. Concerning \textit{adaptive sub-modularity}, we make it an assumption upon which our results is conditioned, and validate it numerically with high precision for small values of $n$ (see Appendix~\ref{sec:appendix-submodular-assumption}).
Direct respective correspondence between our notations and that of \cite{golovin2011adaptive} is given by:
\begin{itemize}
    \item Pool designs $d$ : items $e$;
    \item Test results $T$ : realizations $\Phi$;
    \item Set $\mathcal{D}$ of selected designs : set $E(\pi,\Phi)$ of selected items by policy $\pi$;
    \item $H(p_{S\mid T=t})$ : $f(E(\pi,\Phi), \Phi)$;
    \item $H(S\mid T)$ : $f_{avg}:=\mathbb{E}[f(E(\pi,\Phi), \Phi)]$.
\end{itemize}

This allows one to define, following Definition 1 of \cite{golovin2011adaptive}, the conditional expected marginal benefit of a pool design $d$ given results $t$ as:
\begin{equation}\label{eq:delta}
    \Delta(d):= - [H(S\mid R(S,d)) - H(S)].
\end{equation}
It represents the marginal gain of information obtained, in expectation, by observing the outcome of $d$ at a given stage (this stage being defined by $p_S$, i.e. after having observed test results $t$). \\

\textbf{Adaptive monotonicity} holds if $\Delta(d) \geq 0$ for any $d$.\\

\textbf{Adaptive sub-modularity} holds if for any two sets of results $t$ and $t'$ such that $t$ is a \textit{sub-realization}\footnote{\textit{i.e.} there exist $\mathcal{D}$ and $\mathcal{D}'$ such that $T(S,\mathcal{D})=t$, $T(S,\mathcal{D}')=t'$ and $\mathcal{D}\subset \mathcal{D}'$.} of $t'$, for any pool design $d$: $\Delta(d\mid t)\geq \Delta(d\mid t')$.\\

The below lemma concludes the proof.\\

\textbf{Lemma.} With respect to $\Delta$ defined in Eq.~(\ref{eq:delta}), adaptive monotonicity holds.\\
\textit{Proof.}
Adaptive monotonicity is a consequence of the ``information-never-hurts'' bound $H(X\mid Y)\leq H(X)$ \cite{cover2012elements}.
\begin{flushright}
$\square$
\end{flushright}

\subsection{Theorem~\ref{thm:bloom_info_gain}}

We are interested in the information $I(S,T)$ for a single Bloom filter row with $B$ cells. Because each test in the row contains a disjoint set of patients, $I(S,T)$ is the sum of the information for each test (i.e. the $t_b$ random variables are independent and there are no cross terms).
\begin{align}
    I(S,T) &= H(T) - H(T|S)\\
    &= \sum_{b = 1}^B H(t_b) - \sum_{b = 1}^B H(t_b | S_{\text{patients }\in b})
\end{align}

Using the basic definition of $H(t_b)$, we have that 
\begin{equation}
    H(t_b) = -\sum_{t \in \{0,1\}} \mathrm{Pr}(t_b = t) \log \mathrm{Pr}(t_b = t)
\end{equation}
We use the fact that 
\begin{equation}\mathrm{Pr}(t_b = t) = \sum_{y \in \{0,1\}} \mathrm{Pr}(t_b = t | y_b = y) \mathrm{Pr}(y_b = y)\end{equation}
Since the relationship between the ideal test results $y_b$ and the patient statuses $s_i$ is deterministic, conditioning on $s_i$ is the same as conditioning on $y_b$. In particular, one can write $\mathrm{Pr}(y_b = 0) = \prod (1 - p_i)$. Observe that $tpr = \mathrm{Pr}(t_b = 0 | y_b = 0)$ and $tnr = \mathrm{Pr}(t_b = 1 | y_b = 1)$ and let $\rho_b = \mathrm{Pr}(y_b = 1)$. This gives us a simple expression for $\mathrm{Pr}[t_b = t]$ and thus

\begin{align}
    H(t_b) &= - \left((1 - 2tnr)\rho_b+ tnr\right)\log_2\left((1 - 2tnr)\rho_b+ tnr\right) \\
    &- \left((2tpr - 1) \rho_b + 1 - tpr \right)\log_2\left((2tpr - 1) \rho_b + 1 - tpr \right)
\end{align}

We approach the second term $H(t_b | S_{\text{patients }\in b})$ the same way. 

\begin{equation}
    H(t_b | S_{\text{patients }\in b}) = -\sum_{t\in\{0,1\}}\sum_{y\in\{0,1\}} \mathrm{Pr}[t_b = t, y_b = y] \log \mathrm{Pr}[t_b = t| y_b = y]
\end{equation}

\begin{equation}
    = -\sum_{t\in\{0,1\}}\sum_{y\in\{0,1\}} \mathrm{Pr}[t_b = t| y_b = y]\mathrm{Pr}[y_b = y] \log \mathrm{Pr}[t_b = t| y_b = y]
\end{equation}


\begin{align}
    &= -(1 - \rho_b) (tnr \log_2 (tnr) + (1 - tnr) \log_2 (1 - tnr))\\
    &-\rho_b(tpr\log_2(tpr) + (1 - tpr) \log_2(1-tpr))
\end{align}
Put $\beta = (tnr \log_2 (tnr) + (1 - tnr) \log_2 (1 - tnr))$ and $\alpha = (tpr \log_2 (tpr) + (1 - tpr) \log_2 (1 - tpr))$. Then, the information $I(S,T)$ is equal to
\begin{align}
    I(S,T) &= \sum_{b = 1}^{B}
    - \left((1 - 2tnr)\rho_b+ tnr\right)\log_2\left((1 - 2tnr)\rho_b+ tnr\right) \\
    &- \left((2tpr - 1) \rho_b + 1 - tpr \right)\log_2\left((2tpr - 1) \rho_b + 1 - tpr \right) \\
    &- \left[ - (1 - \rho_b)\beta - \rho_b \alpha\right]
\end{align}

\textbf{Information is Concave in $\rho$:}
To show that there is a single, constant, and optimal probability for each group test to be positive, we prove that $I(S,T)$ is concave in $\rho$. It is sufficient to show that each term $I(S, t_b)$ in the sum is concave in $\rho_b$. 

Taking derivatives, we have 
\begin{align}
    \frac{d}{d\rho_b}I(S,t_b) &= -\frac{1}{\log(2)}(1 - 2tnr) \log((1 - 2tnr) \rho_b + tnr) -\frac{1}{\log(2)} (1 - 2tnr)\\ & -\frac{1}{\log(2)} (2tpr - 1) \log((2tpr - 1)\rho_b + 1 - tpr) -\frac{1}{\log(2)}(2tpr - 1)\\ & - \beta + \alpha
\end{align}

The second derivative is 
\begin{align}
    \frac{d^2}{d\rho_b^2}I(S,t_b) &= -\frac{1}{\log(2)} \left(\frac{(1 - 2tnr)^2 }{(1 - 2tnr)\rho_b + tnr} + \frac{(2tpr - 1)^2}{(2tpr -1)\rho_b - tnr + 1}\right)
\end{align}

We wish to show that $\frac{d^2}{d\rho_b^2}I(S,t_b)\leq 0$, which we will do by proving that the two fractions are both positive. The squared terms in the numerators are positive, as is the expression $(2tpr - 1)\rho_b + 1 - tnr$ because $tnr > 0.5$. This leaves the $(1 - 2tnr)\rho_b + tnr$ term in the denominator. This term is linear in $\rho_b \in [0,1]$, with a minimum of 1 - tnr. Thus, $I(S,T)$ is concave.

\textbf{Optimal Value of $\rho_b$:}
Since the information is concave, there is an optimal value of $\rho_b$ that maximizes the information gain from each grouped test. Since $H(t_b)$ depends only on tpr, tnr and $\rho_b$, it is easy to see that this value is constant and the same for all groups $b$. This proves the theorem. 

However, it is of practical importance to find or approximate the optimal value of $\rho_b$. If one wanted to load balance a variety of (possibly different) priors into groups that have the optimal probability of testing positive, one needs to know the desired value of $\rho_b$. We obtain the following equation by setting the derivative to zero: 
\begin{equation}
    (2tnr - 1) \log((1 - 2tnr)\rho_b + tnr) + (1 - 2tpr) \log( (2tpr - 1)\rho_b + 1 - tpr) = c
\end{equation}
where $c = (1 - 2tnr) + (2tpr -1) + \log(2) (\beta - \alpha)$. One can obtain the optimal $\rho_b$ by numerically solving this equation. When $tpr = tnr$, we have $c = 0$ and the optimal value of $\rho_b = 0.5$.

\subsection{Theorem~\ref{thm:bloom_dimensions}}
We prove the theorem using an analysis that is similar to the one for standard Bloom filters. The Bloom filter decoder identifies a sample as positive if all of the pools containing the sample are positive. It is easy to see that the decoder cannot produce false negatives under perfect tests, because each positive sample will always generate a positive pool result. 
We now analyze the systemic false positives introduced by the pooling operation. Each pool contains either $\frac{N}{B}$ or $\frac{N}{B}-1$ patients, where the latter situation occurs when $B$ does not perfectly divide $N$ and there are a few ``leftover'' elements. Thus, any given sample will share a bin with up to $\frac{N}{B}-1$ other elements, each of which has independent probability $\rho$ of testing positive. To correctly identify a sample as negative, we require that all of these $\frac{N}{B}-1$ samples also test negative. Hence the probability that our sample will not collide with a positive sample is at least 
\begin{equation}
    (1-\rho)^{\frac{N}{B}-1} \leq \exp\left(-\rho (\frac{N}{B}-1)\right)
\end{equation}

The -1 arises from the fact that the sample cannot collide with itself. This analysis holds for a single Bloom filter row, but we have $G$ independent opportunities to land in a negative pool. The rows are independent because independent random hash functions are used to form the groupings. The probability that we collide with a positive in all $G$ groups is at most 
\begin{equation}(1 - (1 - \rho)^{\frac{N}{B}-1})^G\end{equation}
This expression gives the probability that we fail to identify the sample correctly. We want to bound the failure probability $p_f$ and choose parameters that minimize the bound. Note that we replaced $\frac{N}{B}-1$ with $\frac{N}{B}$ - the inequality still holds because $(1 - \rho) < 1$. 

\begin{equation}p_f = (1 - (1-\rho)^{\frac{N}{B}-1})^G \leq \left(1 - \exp\left(-\rho \frac{N}{B}\right)\right)^G\end{equation}

The optimal dimensions for the Bloom filter come from minimizing the upper bound. We use the relation $M = B\times G$ to put $p_f$ in terms of $M$ and $G$. 

\begin{equation}p_f \leq \left(1 - \exp\left(-\rho \frac{N}{M} G\right)\right)^G\end{equation}

We find that the optimal $G = \frac{M}{N \rho} \log 2$. 

\subsection{Theorem~\ref{thm:mitm}}\label{sec:proof-mitm}
\paragraph{Notations.} we use tilda $\tilde{x}$ to denote the estimation of a quantity $x$.

\paragraph{Error Bounds and Confidence Levels.}
Given a test result $t\in\{0,1\}^m$, and a patient $i\in\{1,...,n\}$, we seek to estimate $P[s_i| t]$, \textit{i.e.} the probability of patient sample $s_i$ being positive.  We can rewrite:
\begin{equation}
P[s_i| t] = \dfrac{P[s_i, t]}{P[s_i,  t] +P[\Bar{s_i},t]}=:\dfrac{\lambda}{\lambda+\mu},
\end{equation}
where we defined $\lambda:=P[s_i, t]$ and $\mu:=P[\Bar{s_i}, t]$. Hence, we seek to estimate $\lambda$, resp. $\mu$.
We use the term ``code space'' to refer to the space $\{0,1\}^m$ of encodings of secrets $s\in\{0,1\}^n$. We write $\lambda$ and $\mu$ in terms of the joint distribution of secrets $s$, encodings $c$, and results $t$. Summing across the code space yields:
\begin{equation}
    \lambda = \sum_{c} P[s_i,t,c]= \sum_{c} P[t| s_i,c] P[s_i,c] =\sum_{c} P[t| c] P[s_i,c],
\end{equation}
where the last equality comes from conditional independence of $t$ and $s_i$ w.r.t. $c$.
We now seek to estimate $a(c):=P[t\mid c]$ and $b(c):=P[s_i,c]$. 

Suppose that we have (under-)estimates $\Tilde{a}$ and $\Tilde{b}$ such that $0\leq \max_c ( a(c)-\Tilde{a}(c) ) \leq \varepsilon$ and $0\leq \sum_c ( b(c)-\Tilde{b}(c) ) \leq \varepsilon$. Later, we will describe how to obtain these estimates. For now, observe that we can (under-)estimate $\lambda=\sum_c a(c)b(c)$ with $\Tilde{\lambda}:=\sum_c \Tilde{a}(c)\Tilde{b}(c)$, with the following error bound\footnote{Note that for any $c$, we have: $a(c),b(c)\leq 1$, and that $\tilde{a}(c) \leq a(c)$ and $\tilde{b}(c) \leq b(c)$ because they are under-estimates.}:
\begin{align}
0 \leq    \lambda - \Tilde{\lambda} &= \sum_c a(c)b(c) - \sum_c \Tilde{a}(c)\Tilde{b}(c)\\
&=\sum_c a(c)(b(c)-\Tilde{b}(c)) + \sum_c (a(c)-\Tilde{a}(c))\Tilde{b}(c)\\
&\leq \sum_c (b(c)-\Tilde{b}(c)) + \max_c (a(c)-\Tilde{a}(c))\\
&\leq 2\varepsilon,
\end{align}
and similarly $0\leq \mu-\Tilde{\mu}\leq 2\varepsilon$, which would imply $0\leq (\lambda+\mu)-(\tilde{\lambda}+\tilde{\mu})\leq 4\varepsilon$; however, we can obtain a tighter upper bound of $3\varepsilon$ by noticing that $\sum_{c} P[s_i,c]+\sum_{c}P[\Bar{s_i},c]=\sum_c P[c]\leq 1$, yielding a true $P[s_i|t]$ in the below (arithmetic) interval:
\begin{equation}
    P[s_i|t]\in  [\Tilde{\lambda}, \Tilde{\lambda} + 2\varepsilon] / [\Tilde{\lambda} +\Tilde{\mu},\Tilde{\lambda} +\Tilde{\mu} + 3\varepsilon] =  \left[\dfrac{\Tilde{\lambda}}{\Tilde{\lambda}+\Tilde{\mu}+3\varepsilon}, \dfrac{\Tilde{\lambda}+2\varepsilon}{\Tilde{\lambda}+\Tilde{\mu}}  \right].
\end{equation}
We want to bound the size of this interval to show that our estimate is close to the true $P[s_i|t]$. We do this via a Taylor alternate series:
\begin{align}
    \dfrac{\Tilde{\lambda}+2\varepsilon}{\Tilde{\lambda}+\Tilde{\mu}} - \dfrac{\Tilde{\lambda}}{\Tilde{\lambda}+\Tilde{\mu}+3\varepsilon} &\leq \dfrac{2\varepsilon}{\Tilde{\lambda} + \Tilde{\mu}} + \dfrac{3\varepsilon\Tilde{\lambda}}{(\Tilde{\lambda}+\Tilde{\mu})^2}\\
    &= \varepsilon \dfrac{5\Tilde{\lambda} + 2\Tilde{\mu}}{(\Tilde{\lambda}+\Tilde{\mu})^2}\\
    &\leq \dfrac{5\varepsilon}{\Tilde{\lambda}+\Tilde{\mu}},
\end{align}
which concludes the proof that we can estimate $P[s_i|t]$ with error less than $5\varepsilon/\Tilde{P}[t]$, where $\Tilde{P}[t]:=\Tilde{\lambda}+\Tilde{\mu}$, given estimates $\tilde{a}$ and $\tilde{b}$. Hence, we only need to construct the under-estimates $\Tilde{a}$ and $\Tilde{b}$ such that $0\leq \max_c a(c)-\Tilde{a}(c) \leq \varepsilon$ and $0\leq \sum_c b(c)-\Tilde{b}(c) \leq \varepsilon$. To construct $\tilde{b}(c)$, assume we have an integer $k$ such that $f(k):=\sum_{j=k}^n {n \choose j} p^j (1-p)^{n-j} < \varepsilon$. Let
\begin{equation}
\label{eq:b_tilde_formula}
    \Tilde{b}(c) := \sum_{\substack{s\in\{0,1\}^n\\ \sum_j s_j < k\\ enc(s)=c }} P[s_i,c].
\end{equation}
Then, 
\begin{align}
    \sum_c b(c)-\Tilde{b}(c) &\leq  \sum_c\sum_{\substack{s\in\{0,1\}^n\\ \sum_j s_j \geq k} } P[s_i,c]\\
    &\leq \sum_{\substack{s\in\{0,1\}^n\\ \sum_j s_j \geq k }} P[s_i]\\
    &= f(k)\\
    &\leq \varepsilon.
\end{align}
Similarly, let
\begin{equation}
    \Tilde{a}(c) := \mathbf{1}_{\{prob[FP][FN]>\varepsilon\}} prob[FP][FN]
\end{equation}
where $\mathbf{1}$ is the indicator function. The $prob[FP][FN]$ term is the probability $P[t|c]$ of getting a particular (corrupted) output $t$ given a (true) code $c$. This term is defined as follows, where $FP,FN,N,P$ are the number of false positives $FP$, false negatives $FN$, total negatives $N$ and total positives $P$ in the output $t$ when compared with $c$. Note that $N+P = m$ and that $prob[FP][FN] = a(c)$. We use the term $prob[FP][FN]$ only for \textbf{notational convenience} to show that $a(c)$ depends on $FP,FN,N$ and $P$. 

\begin{equation}\label{eq:probfpfn}
    prob[FP][FN]:= (1-\mathrm{tnr})^{FP}\mathrm{\mathrm{tpr}}^{P-FP}(1-\mathrm{\mathrm{tpr}})^{FN}\mathrm{\mathrm{tnr}}^{N-FN} = a(c).
\end{equation}
Then, 
\begin{align}
    a(c)-\Tilde{a}(c)&= prob[FP][FN] - \mathbf{1}_{\{prob[FP][FN] > \varepsilon\}} prob[FP][FN]\\
    &\leq \mathbf{1}_{\{prob[FP][FN] \leq \varepsilon\}} prob[FP][FN]\\
    &\leq \varepsilon.
\end{align}

This concludes our presentation of the estimators $\tilde{a}(c)$ and $\tilde{b}(c)$. Note that we presented a confidence interval together with a bound on its size, \textit{i.e.} we showed that the true value $P[s_i|t]$ is within an interval that depends on the observed quantity $\tilde{P}[s_i|t]$. However, we can also provide an interval for the observed quantity as a function of the true value:
\begin{equation}
    \Tilde{P}[s_i|t]\in  [\lambda -2\varepsilon, \lambda] / [\lambda +\mu - 3\varepsilon,\lambda +\mu] =  \left[\dfrac{\lambda-2\varepsilon}{\lambda+\mu}, \dfrac{\lambda}{\lambda+\mu- 3\varepsilon}  \right],
\end{equation}
whose size can be bounded by:
\begin{align}
    \dfrac{\lambda}{\lambda+\mu- 3\varepsilon} -\dfrac{\lambda-2\varepsilon}{\lambda+\mu} &= \dfrac{3\varepsilon \lambda}{(\lambda+\mu)^2}+ \dfrac{2\varepsilon}{\lambda+\mu} +\dfrac{3\lambda\varepsilon^2}{(\lambda+\mu)^3}\sum_{j=0}^{+\infty} \left(\dfrac{\varepsilon}{\lambda+\mu}\right)^j \\
    &\leq  \dfrac{5\varepsilon}{\lambda+\mu} + \dfrac{3\lambda\varepsilon^2}{(\lambda+\mu)^3}\dfrac{1}{1-\frac{\varepsilon}{\lambda+\mu}} \\
    &\leq  \dfrac{5\varepsilon}{\lambda+\mu} + \dfrac{\varepsilon}{\lambda+\mu}\\
    &= \dfrac{6\varepsilon}{\lambda+\mu}\\
    &=\dfrac{6\varepsilon}{P[t]}, 
\end{align}
where we assumed $\varepsilon <\lambda /4$ to justify that $\frac{3\lambda\varepsilon^2}{(\lambda+\mu)^3}\frac{1}{1-\frac{\varepsilon}{\lambda+\mu}} < \frac{\varepsilon}{\lambda+\mu}$. One might also be interested in the error rather than a confidence interval. Recall that $\tilde{P}[t]:=\tilde{\lambda}+\tilde{\mu}\leq \lambda+\mu = P[t]$
If we want the error of our estimator, then one can easily show that $|\tilde{P}[s_i|t]-P[s_i|t]|\leq 4\varepsilon /P[t] \leq 4\varepsilon /\tilde{P}[t]$. In practice, $\tilde{P}[t]$ can be computed to get upper bounds on the estimation error and confidence level.


We will now present an algorithm that efficiently computes these estimators. In our algorithm, $A(\epsilon)$ is the number of secrets with at most $k$ nonzeros, $C(\epsilon)$ is number of codes produced by this restricted set of $k$-sparse secrets, and $B(\epsilon)$ is a set of probable ideal codes for the potentially-corrupted output $t$ that we observe.  

\begin{algorithm}[H]
\SetAlgoLined
\SetKwInOut{Input}{Input}\SetKwInOut{Output}{Output}
\Input{$n$ \& $m$, $\mathrm{\mathrm{tpr}}$ \& $\mathrm{\mathrm{tnr}}$, prevalence $p$, test results $t\in\{0,1\}^m$, precision parameter $\varepsilon$;}
\Output{Estimates $\tilde{P}[s_i|t]$ for $i\in\{1,...,n\}$ with $|\tilde{P}[s_i|t]-P[s_i|t]|\leq 4\varepsilon /P[t]$;}
\BlankLine
\textbf{Preprocessing:} (independent of results $t$)\\
Compute $k$ such that $f(k)<\varepsilon$ and initialize $\mathcal{C}=\emptyset$\;
Enumerate all the codes $c:=enc(s)$ for $s$ with less than $k$ positives\footnote{This yields a set of size $C(\varepsilon)$ computed in time $A(\varepsilon)$.}\;
Use these codes to approximate $\tilde{P}[s_i,c]$ and $\tilde{P}[\Bar{s_i},c]$ using the formula for $\tilde{b}(c)$ in Eq.~(\ref{eq:b_tilde_formula}). Store the results in $\mathcal{C}$\;
\textbf{Query:} (dependent upon results $t$)\\
Compute $P:=\sum_i t_i$, $N:=m-P$ and $a(c)$ (see Eq.~(\ref{eq:probfpfn})) for $FP\leq P$, $FN\leq N$\;
Compute $B(\varepsilon) := \sum_{a(c)>\varepsilon} {P \choose FP}{N \choose FN}$\;
\eIf{$C(\varepsilon) < B(\varepsilon)$}
{Estimate $P[s_i,t]$ (resp. $P[\Bar{s_i},t]$) by iterating over the codes $c$ in $\mathcal{C}$ and reporting $\sum_c a(c) \Tilde{P}[s_i,c]$\;}
{Enumerate\footnote{We can recursively enumerate these codes in time $B(\varepsilon)$ since $a(c)$ is monotonic w.r.t. both variables, by starting the enumeration at $c:=t$, \textit{i.e.} $FP=FN=0$, and recursively increment $FP$ or $FN$.} codes $c$ such that $a(c)>\varepsilon$\;}
Output final estimates $\tilde{P}[s_i|t]:=\tilde{P}[s_i,t]/(\tilde{P}[s_i,t]+\tilde{P}[\Bar{s_i},t])$\;
 \caption{(MITM Decoder)}\label{alg:mitm-decoder}
\end{algorithm}

\paragraph{Complexity Analysis.} Since the outcome of a test $t$ is conditionally independent to $s$ w.r.t. $c$, we can pre-compute all encodings $c:=enc(s)$ for $s$ belonging to the reduced search space of size $A(\varepsilon):=\sum_{i=0}^{k-1}{n \choose i}$. Saving all these resulting encodings with a hashmap or a set structure gives a space of complexity proportional to $C(\varepsilon)\leq A(\varepsilon)$, since the output function image of an input set is always smaller than (or equal to) the size of the input set. Finally, at query time, we seek to estimate $P[s_i|t]$. Note that we have pruned two search spaces: the space of encodings of $A(\varepsilon)$ many secrets, reduced from $2^m$ to $C(\varepsilon)$, and the space of codes $c$ such that for our given $t$, $prob[FP][FN]>\varepsilon$, reduced from $2^m$ to $B(\varepsilon)$. Given a test result $t$, we can compute $N,P$ in $\mathcal{O}(m)$ operations, which then allows us to compute $B(\varepsilon)$ for this $t$. Also note that we approximate $P[s_i,t]$ via $\sum_{c} \tilde{a}(c)\tilde{b}(c)$. Since $\tilde{a}(c)=0$ for $c$ such that $t$ doesn't belong to the reduced test results space of size $B(\varepsilon)$, we can choose to perform this sum on either this set, or the reduced code space of size $C(\varepsilon)$: whichever is the smallest. This is where the denomination ``meet-in-the-middle'' comes from.

\section{Interactive demonstration}

The C++ code can be used in the browser through an interactive WebAssembly demo:\\
\href{https://bloom-origami.github.io/}{https://bloom-origami.github.io/}

The following features are implemented:

\begin{itemize}
    \item Bloom assay generation
    \item Greedy adaptive strategy simulation
    \item Design optimization using genetic algorithms
    \item Posterior decoding using MITM 
\end{itemize}

\iftrue
\section{Prevalence Estimation}

Our designs assume that the prevalence $\rho$ is known, at least approximately. However, we can also use our Bloom filter design to estimate the prevalence in the overall infected population. When we randomly and independently sample an individual from the population, they have probability $\rho$ of being infected. The prevalence estimation problem is to determine $\rho$ using as few tests as possible. Here, we assume perfect tests to simplify the analysis. 

Of course, one could individually test a large number of people from the population and report the fraction of positive test results. The challenge is that if we screen individuals, we end up with a random variable for which the mean to variance ratio is unfavorable. Consider a random variable $X \in \{0, 1\}$ with $\mathbb{E}[X] = \rho$ and variance $\mathrm{var}[X] = \rho - \rho^2 = \rho(1-\rho)$. The error of the empirical average of $m$ individual tests is
$$\frac{1}{m} \sum_{i=1}^m X_i - \mathbb{E}[X] = O\left(\frac{\mathrm{std}[X_u]}{\sqrt{m}}\right).$$
The relative error is $\sqrt{\frac{1-\rho}{\rho}}$. Clearly this is minimized for $\rho = 1$. Unfortunately, this value is entirely useless since it corresponds to the situation where every test returns positive. In practice, we encounter the unfortunate situation of $\rho << 1$ where the relative error diverges. Under a naive random sampling approach to prevalence estimation, a very large number of tests are required. To amend this situation, it is beneficial to increase the probability of a positive test by testing multiple candidates at once. Our pooled tests are no longer positive with probability $\rho$ but with probability $q = 1 - (1-\rho)^k$, where $k$ is the number of samples combined in a single pool. Knowing $q$, we can solve for $\rho$ via

$$\rho = 1 - (1 - q)^\frac{1}{k}$$


We will use the central limit theorem and the delta method to show that we need fewer Bloom filter pooled tests than random individual tests to estimate the prevalence. The central limit theorem states that

$$ \overline{X}_m - \mu \overset{d}{\to} \mathcal{N}\left(0,\frac{\sigma(\mu)^2}{m}\right)$$

where $\overline{X}_m$ is the average of $m$ trials, $\mu = \mathbb{E}[X]$, and $\sigma^2 = \mathrm{var}[X]$. The delta method states that if we have a function $g(x)$ and its derivative $g'(x)$, then 

$$ g(\overline{X}_m) - g(\mu) \overset{d}{\to} \mathcal{N}\left(0,\frac{[g'(\mu)\sigma(\mu)]^2}{m}\right)$$

\subsection{Prevalence Estimation with Random Sampling}

Suppose we randomly sample individuals from the population and perform $m$ individual tests. Here, $X$ is the test status of the patient and it is positive with prevalence $\rho$. We estimate $\rho$ as $\hat{\rho} = \frac{1}{m}\sum_{i = 1}^m X_i$
Observe that $\mathbb{E}[X] = \mu = \rho$ and $\mathrm{var}[X] = \rho(1 - \rho)$. Use the central limit theorem to observe that

$$ \hat{\rho} - \rho \overset{d}{\to} \mathcal{N}\left(0,\frac{\rho(1 - \rho)}{m}\right) $$

\subsection{Prevalence Estimation with Bloom Filters}

Suppose we combine $k$ samples into each bin. Now $X$ is the test status of the bin and it is positive with probability $q = 1 - (1 - \rho)^k$. Hence $\mu = q$ and $\sigma^2 = q(1 - q)$. Use the delta method with

$$g(x) = 1 - (1 - x)^{\frac{1}{k}}$$
$$g'(x) = \frac{1}{k}(1 - x)^{\frac{1}{k} - 1}$$

Observe that $g(\mu) = \rho$ and that $\hat{\rho} = g(\overline{X}_n)$. From the delta theorem we have 

$$ \hat{\rho} - \rho \overset{d}{\to} \mathcal{N}\left(0,\frac{[g'(\mu)\sigma(\mu)]^2}{m}\right)$$

We proceed by analyzing the $g'(\mu)\sigma(\mu)$ term. This term is 
$$ \sqrt{\mu (1 - \mu)} \frac{1}{k} (1 - \mu)^{\frac{1}{k} - 1}$$
Recall that $\mu = q = 1 - (1 - \rho)^k$. Substitute this value to get

$$ \hat{\rho} - \rho \overset{d}{\to} \mathcal{N}\left(0,\frac{\alpha^2}{m}\right)$$
where 
$$ \alpha = \frac{1}{k} \frac{\sqrt{1 - (1 - \rho)^k}}{(1 - \rho)^{\frac{k}{2}-1}}$$

\subsection{Comparison}

We are interested in whether the variance of the Bloom filter estimator is larger than the variance of the random sampling estimator. That is, we want to prove the following inequality. 

$$ \frac{1}{k} \frac{\sqrt{1 - (1 - \rho)^k}}{(1 - \rho)^{\frac{k}{2}-1}} \leq \sqrt{\rho(1 - p)}$$
Rearrange
$$ \sqrt{\frac{1}{(1 - \rho)^k} - 1} \left(\frac{1 - \rho}{k}\right) \leq \sqrt{\rho(1 - \rho)}$$

Recall the inequality $1 - x \geq e^{-x\,/\,(1-x)}$ when $0 \geq x < 1$. Applied to our situation, this means that 
$$\frac{1}{(1 - \rho)^k} < e^{\rho k \,/\, (1 - \rho)}$$ Therefore our inequality becomes
$$ \sqrt{e^{\frac{\rho k}{1-\rho}} - 1} \left(\frac{1-\rho}{k}\right) \leq \sqrt{\rho(1 - \rho)}$$
Put $k = (1 - \rho) / \rho$. Then the inequality is true when $\rho \leq 1/e$. Bloom filters are a better way to measure prevalence provided that $\rho$ is smaller than 37\% or (using symmetry arguments) greater than 63\%.

\fi

\section{Empirical Validation of Adaptive Sub-Modularity}\label{sec:appendix-submodular-assumption}

Below the C++ code used to validate the assumption of adaptive sub-modularity relative to Theorem~\ref{thm:greedy}, for small values of $n$.

\begin{lstlisting}[language=C++, caption={}]

#include <vector>
#include <algorithm>
#include <utility>
#include <math.h>
#include <assert.h>
#include <iostream>
#include <functional>
#include <map>
#include <queue>

using namespace std;

using vd =  vector<double>;

// expected entropy of simultaneous tests
double expected_entropy(const double obs01, const double obs11,
                        const vd &prior,
                        const vector<int> &tests) {
    // optimized version with constant memory
    int t = tests.size()   ;
    int N = prior.size();
    double ans = 0;
    for(int m=0; m<1<<t; m++) {
        double prob_m = 0;
        double entropy_m = 0;
        for(int s=0; s<N; s++) {
            double joint_s_m = prior[s];
            // probability of observing joint_s_m
            for(int i=0; i<t; i++) {
                auto p = (s & tests[i]) ? obs11 : obs01;
                joint_s_m *= (m & (1<<i)) ? p : 1-p;
            }
            prob_m += joint_s_m;
            if(joint_s_m)
                entropy_m -= joint_s_m * log2(joint_s_m);
        }
        if(prob_m)
            entropy_m += prob_m * log2(prob_m);
        ans +=  entropy_m;
    }
    return ans;
}

static double drand() {
    return (double)rand() / RAND_MAX;
}

int main() {
    int TESTS = 100000;
    while(TESTS--) {

        int n = 5;
        int N = 1 << n;
        double obs01 = drand() / 2;
        double obs11 = 1 - drand() / 2;

//        vd prob_ill(n);
//        for(auto &v : prob_ill)
//            v = drand();
//        auto prior = factor(prob_ill);
        vd prior(N);
        double s = 0;
        for(auto &v : prior)
            s += v = drand();
        for(auto &v : prior)
            v /= s;

        int test1 = rand() % (N-1) + 1;
        int test2 = rand() % (N-1) + 1;

        auto aux = [&](const vector<int> &tests) {
            return expected_entropy(obs01, obs11, prior, tests);
        };

        auto delta = aux({test1, test2}) - aux({test1}) 
            - aux({test2}) + aux({});

        if(delta < -1e-6) {
            cout << test1 << ' ' << test2 << endl;
            cout << obs01 << ' ' << obs11 << endl;
//            for(auto x : prob_ill)
//                cout << x << ' ';
            cout << endl;

            cout << delta << endl;
            cout << aux({}) << ' ' << aux({test1}) << ' ' 
                << aux({test2}) << ' ' << aux({test1, test2}) << endl;
        }
    }
}
static double drand() {
    return (double)rand() / RAND_MAX;
}
int main() {
    int TESTS = 100000;
    while(TESTS--) {
        int n = 5;
        int N = 1 << n;
        double obs01 = drand() / 2;
        double obs11 = 1 - drand() / 2;
//        vd prob_ill(n);
//        for(auto &v : prob_ill)
//            v = drand();
//        auto prior = factor(prob_ill);
        vd prior(N);
        double s = 0;
        for(auto &v : prior)
            s += v = drand();
        for(auto &v : prior)
            v /= s;

        int test1 = rand() % (N-1) + 1;
        int test2 = rand() % (N-1) + 1;
        auto aux = [&](const vector<int> &tests) {
            return expected_entropy(obs01, obs11, prior, tests);
        };
        auto delta = aux({test1, test2}) 
            - aux({test1}) - aux({test2}) + aux({});
        if(delta < -1e-6) {
            cout << test1 << ' ' << test2 << endl;
            cout << obs01 << ' ' << obs11 << endl;
//            for(auto x : prob_ill)
//                cout << x << ' ';
            cout << endl;
            cout << delta << endl;
            cout << aux({}) << ' ' << aux({test1}) << ' ' 
                << aux({test2}) << ' ' << aux({test1, test2}) << endl;
        }
    }
}

\end{lstlisting}

%% file: main.bbl
\begin{thebibliography}{10}

\bibitem{Bloom}
Burton~H. Bloom.
\newblock Space/time trade-offs in hash coding with allowable errors.
\newblock {\em Commun. ACM}, 13(7):422–426, 1970.

\bibitem{broder2020note}
Andrei~Z Broder and Ravi Kumar.
\newblock A note on double pooling tests.
\newblock {\em arXiv preprint arXiv:2004.01684}, 2020.

\bibitem{chan2014non}
Chun~Lam Chan, Sidharth Jaggi, Venkatesh Saligrama, and Samar Agnihotri.
\newblock Non-adaptive group testing: Explicit bounds and novel algorithms.
\newblock {\em IEEE Transactions on Information Theory}, 60(5):3019--3035,
  2014.

\bibitem{cheraghchi2012graph}
Mahdi Cheraghchi, Amin Karbasi, Soheil Mohajer, and Venkatesh Saligrama.
\newblock Graph-constrained group testing.
\newblock {\em IEEE Transactions on Information Theory}, 58(1):248--262, 2012.

\bibitem{cover2012elements}
Thomas~M Cover and Joy~A Thomas.
\newblock {\em Elements of information theory}.
\newblock John Wiley \& Sons, 2012.

\bibitem{golovin2011adaptive}
Daniel Golovin and Andreas Krause.
\newblock Adaptive submodularity: Theory and applications in active learning
  and stochastic optimization.
\newblock {\em Journal of Artificial Intelligence Research}, 42:427--486, 2011.

\bibitem{guestrin2005near}
Carlos Guestrin, Andreas Krause, and Ajit~Paul Singh.
\newblock Near-optimal sensor placements in gaussian processes.
\newblock In {\em Proceedings of the 22nd international conference on Machine
  learning}, pages 265--272, 2005.

\bibitem{he2020temporal}
Xi~He, Eric H.~Y. Lau, Peng Wu, Xilong Deng, Jian Wang, Xinxin Hao, Yiu~Chung
  Lau, Jessica~Y. Wong, Yujuan Guan, Xinghua Tan, Xiaoneng Mo, Yanqing Chen,
  Baolin Liao, Weilie Chen, Fengyu Hu, Qing Zhang, Mingqiu Zhong, Yanrong Wu,
  Lingzhai Zhao, Fuchun Zhang, Benjamin~J. Cowling, Fang Li, and Gabriel~M.
  Leung.
\newblock Temporal dynamics in viral shedding and transmissibility of
  {COVID}-19.
\newblock {\em Nature Medicine}, 26(5):672--675, 2020.

\bibitem{hwang1987non}
FK~Hwang and VT~S{\'o}s.
\newblock Non-adaptive hypergeometric group testing.
\newblock {\em Studia Sci. Math. Hungar}, 22(1-4):257--263, 1987.

\bibitem{indyk2010efficiently}
Piotr Indyk, Hung~Q Ngo, and Atri Rudra.
\newblock Efficiently decodable non-adaptive group testing.
\newblock In {\em Proceedings of the twenty-first annual ACM-SIAM symposium on
  Discrete Algorithms}, pages 1126--1142. SIAM, 2010.

\bibitem{jaakkola1999variational}
Tommi~S Jaakkola and Michael~I Jordan.
\newblock Variational probabilistic inference and the qmr-dt network.
\newblock {\em Journal of artificial intelligence research}, 10:291--322, 1999.

\bibitem{gitCovidBloom}
Tomas Janousek.
\newblock https://github.com/liskin/covid19-bloom.
\newblock \url{https://github.com/liskin/covid19-bloom}, 2020.

\bibitem{knill1998non}
Emanuel Knill, William~J Bruno, and David~C Torney.
\newblock Non-adaptive group testing in the presence of errors.
\newblock {\em Discrete applied mathematics}, 88(1-3):261--290, 1998.

\bibitem{koller2009probabilistic}
Daphne Koller and Nir Friedman.
\newblock {\em Probabilistic graphical models: principles and techniques}.
\newblock MIT press, 2009.

\bibitem{Smola}
Stefan Lohse, Thorsten Pfuhl, Barbara Berkó-Göttel, Jürgen Rissland, Tobias
  Geißler, Barbara Gärtner, Sören~L Becker, Sophie Schneitler, and Sigrun
  Smola.
\newblock Pooling of samples for testing for {SARS-CoV-2} in asymptomatic
  people.
\newblock {\em The Lancet Infectious Diseases}, 2020.
\newblock https://doi.org/10.1016/S1473-3099(20)30362-5.

\bibitem{lu2008counter}
Yi~Lu, Andrea Montanari, Balaji Prabhakar, Sarang Dharmapurikar, and Abdul
  Kabbani.
\newblock Counter braids: a novel counter architecture for per-flow
  measurement.
\newblock {\em ACM SIGMETRICS Performance Evaluation Review}, 36(1):121--132,
  2008.

\bibitem{luby1993optimal}
Michael Luby, Alistair Sinclair, and David Zuckerman.
\newblock Optimal speedup of {Las Vegas} algorithms.
\newblock {\em Information Processing Letters}, 47(4):173--180, 1993.

\bibitem{mazumdar2016nonadaptive}
Arya Mazumdar.
\newblock Nonadaptive group testing with random set of defectives.
\newblock {\em IEEE Transactions on Information Theory}, 62(12):7522--7531,
  2016.

\bibitem{mich2020bloom}
Monika Mich~Cechova.
\newblock Bloom-filter inspired testing of pooled samples (and splitting of
  swabs!).
\newblock April 1, 2020.

\bibitem{mitzenmacher2017probability}
Michael Mitzenmacher and Eli Upfal.
\newblock {\em Probability and computing: Randomization and probabilistic
  techniques in algorithms and data analysis}.
\newblock Cambridge university press, 2017.

\bibitem{Padhye2020.04.24.20078949}
Nikhil~S Padhye.
\newblock Reconstructed diagnostic sensitivity and specificity of the rt-pcr
  test for covid-19.
\newblock {\em medRxiv}, 2020.

\bibitem{Schmidt2020.04.28.20074187}
Michael Schmidt, Sebastian Hoehl, Annemarie Berger, Heinz Zeichhardt, Kai
  Hourfar, Sandra Ciesek, and Erhard Seifried.
\newblock {FACT} - {F}rankfurt adjusted {COVID-19} testing - a novel method
  enables high-throughput {SARS-CoV-2} screening without loss of sensitivity.
\newblock {\em medRxiv}, 2020.

\bibitem{inflammatory}
Matthew~Zirui Tay, Chek~Meng Poh, Laurent R{\'e}nia, Paul~A. MacAry, and Lisa
  F.~P. Ng.
\newblock The trinity of {COVID-19}: immunity, inflammation and intervention.
\newblock {\em Nature Reviews Immunology}, 20(6):363--374, 2020.

\bibitem{vaismanmodel}
Radislav Vaisman, Ofer Strichman, and Ilya Gertsbakh.
\newblock Model counting of monotone cnf formulas with spectra.

\bibitem{origami}
Gaolian Xu, Debbie Nolder, Julien Reboud, Mary Oguike, Donelly van Schalkwyk,
  Colin Sutherland, and Jonathan Cooper.
\newblock Paper‐origami‐based multiplexed malaria diagnostics from whole
  blood.
\newblock {\em Angewandte Chemie International Edition}, 55, 08 2016.

\bibitem{evaluation2020yelin}
Idan Yelin, Noga Aharony, Einat Shaer-Tamar, Amir Argoetti, Esther Messer, Dina
  Berenbaum, Einat Shafran, Areen Kuzli, Nagam Gandali, Tamar Hashimshony, Yael
  Mandel-Gutfreund, Michael Halberthal, Yuval Geffen, Moran Szwarcwort-Cohen,
  and Roy Kishony.
\newblock Evaluation of {COVID-19} {RT-qPCR} test in multi-sample pools.
\newblock {\em Clinical Infectious Diseases}, 2020.

\end{thebibliography}
